\pdfoutput=1
\documentclass[letterpaper, 10 pt, conference]{ieeeconf}  

\IEEEoverridecommandlockouts                              

\usepackage{graphics} 
\usepackage{amsmath} 
\usepackage{amssymb}  
\usepackage{array}
\usepackage{multirow}
\usepackage{times}
\usepackage{epsfig}
\usepackage{graphicx}
\usepackage{color}
\usepackage{booktabs} 
\usepackage{pifont}

\usepackage{hyperref}

\title{\LARGE \bf
Real-time CPU-based large-scale 3D mesh reconstruction
}

\newcommand{\cmark}{\ding{51}}%
\newcommand{\xmark}{}%
\author{Enrico Piazza$^{1}$  Andrea Romanoni$^{1}$  Matteo Matteucci$^{1}$
\thanks{$^{1}${Politecnico di Milano, Dipartimento di Elettronica, Informazione e Bioingegneria (DEIB),  Milano, Italy}
        \vfill{\tt\scriptsize enrico.piazza@polimi.it}
		\vfill{\tt\scriptsize andrea.romanoni@polimi.it} 
        \vfill{\tt\scriptsize matteo.matteucci@polimi.it (corresponding author)}
        }
}

\begin{document}

\maketitle
\thispagestyle{empty}
\pagestyle{empty}

\begin{abstract}
In Robotics, especially in this era of autonomous driving, mapping is one key ability of a robot to be able to navigate through an environment, localize on it and analyze its traversability.
To allow for real-time execution on constrained hardware, the map usually estimated by feature-based or semi-dense SLAM algorithms is a sparse point cloud; a richer and more complete representation of the environment is desirable. 
Existing dense mapping algorithms require extensive use of GPU computing and they hardly scale to large environments; incremental algorithms from sparse points still represent an effective solution when light computational effort is needed and big sequences have to be processed in real-time. 
In this paper we improved and extended the state of the art incremental manifold mesh algorithm proposed in \cite{litvinov_lhuillier_13} and extended in \cite{romanoni15b}. 
While these algorithms do not achieve real-time and they embed points from SLAM or Structure from Motion only when their position is fixed, in this paper we propose the first incremental algorithm able to reconstruct a manifold mesh in real-time through single core CPU processing which is also able to modify the mesh according to 3D points updates from the underlying SLAM algorithm.
We tested our algorithm against two state of the art incremental mesh mapping systems on the KITTI dataset, and we showed that, while accuracy is comparable, our approach is able to reach real-time performances thanks to an order of magnitude speed-up.
\end{abstract}

\section{INTRODUCTION}
Robot navigation and localization is a long studied topic in Robotics. 
In the last decade, the advancements in Simultaneous Localization and Mapping (SLAM) algorithms, especially since the proposal of the Parallel Tracking and Mapping (PTAM) \cite{klein_murray07} paradigm, have led to algorithms results in robot localization and environment mapping. 
The most preeminent paradigms shown in the literature can be classified in feature-based~\cite{davison2007monoslam,strasdat11},  direct~\cite{engel2014lsd,engel2015large}, and dense algorithms~\cite{newcombe2011dtam,kahler2016real}.

Feature-based and direct approaches are able to reconstruct a sparse or semi-dense map of large-scale environment on a CPU
On the other hand, even if dense algorithms have been proved to be scalable, as in \cite{whelan2012kintinuous}, they rely on  GPU-computing.
However, in many robotics applications, such as autonomous vehicle and cost-effective surveying, computational power is a limited resource and, when available, GPUs are not as powerful as needed by dense mapping algorithms. 
%

To propose a trade-off between accuracy and computation effort and to have a dense output, some works in literature have recently focused on incremental reconstruction of a mesh from sparse feature point extracted by SLAM algorithms~\cite{lovi_et_al_11,litvinov_lhuillier_13,romanoni15b}. 
These methods estimate a low resolution model of the environment, represented by a dense mesh, which is informative enough for traversability analysis and path planning.
Among these methods, the most advanced enforce the manifold property\footnote{A mesh is manifold if each vertex v is regular, i.e., if and only if the
edges connecting the vertices opposite to v are homeomorphic to a disk .i.e., they form  path without loops and closed (see \cite{lhuillier20152} for more details).} 
which is needed for both online improvement, e.g., in mesh smoothing~\cite{taubin1995signal}, or offline refinements, e.g., in photometric mesh optimization~\cite{vu_et_al_2012,romanoni16,romanoni17,romanoni2017multi}. This property can be enforced incrementally as in~\cite{litvinov_lhuillier_13} and~\cite{romanoni15b}, but this makes the existing incremental  algorithms not able to run in real-time.

\begin{figure}[tp]
 \centering
 \begin{tabular}{c}
  \hspace{-0.25cm}\includegraphics[width=\columnwidth]{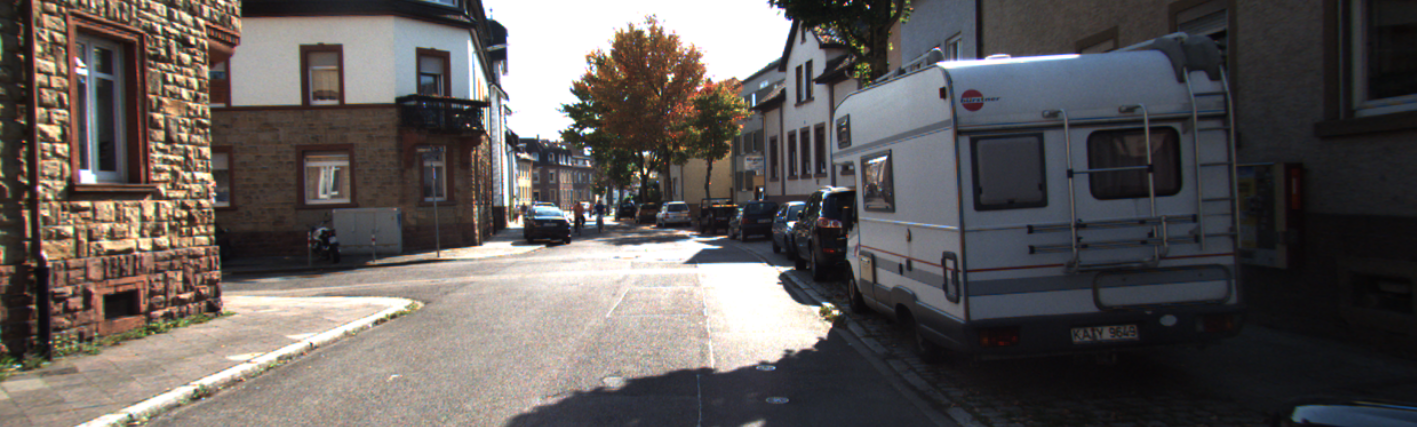}\\
  \hspace{-0.25cm}\includegraphics[width=\columnwidth]{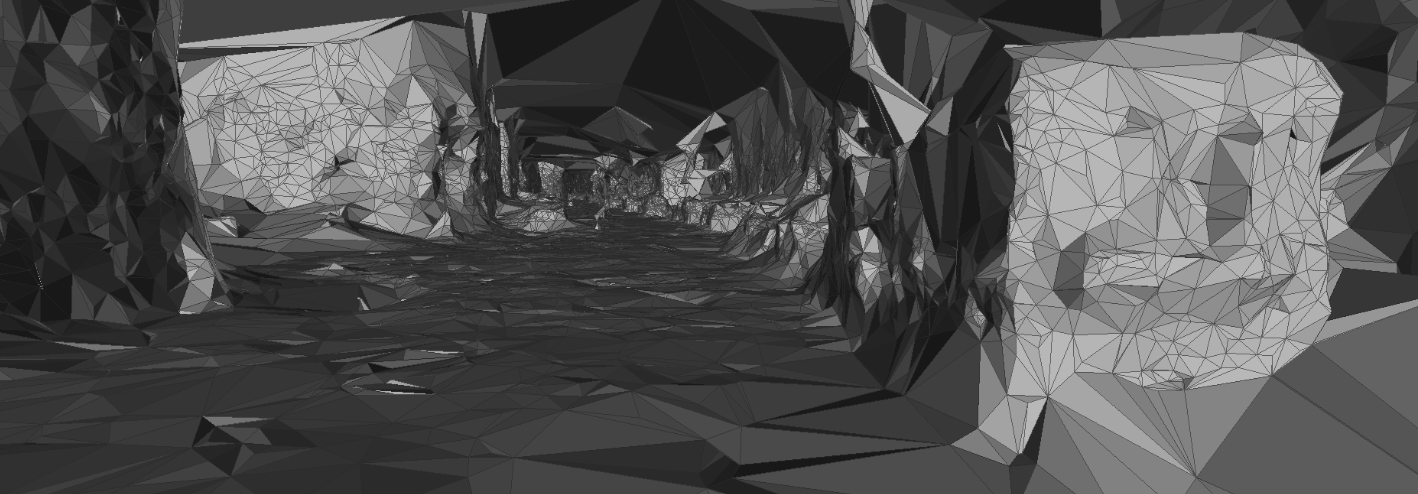}\\
 \end{tabular}
 \caption{Mesh reconstruction in real-time with the proposed algorithm.}
 \label{fig:exRec}
\end{figure}

Starting from the previous works on incremental mesh reconstruction~\cite{romanoni15a,romanoni15b}, in this paper we propose the first real-time incremental algorithm that is able to reconstruct a manifold mesh of the scene running on a single CPU core, and leaving other cores available to perform camera tracking and sparse data estimation via classical feature-based SLAM.
As a  main contribution we improve the computational effort of the methods proposed in~\cite{litvinov_lhuillier_13} and~\cite{romanoni15b}, reaching a comparable accuracy (see in Figure~\ref{fig:exRec} an example of result).
An open source implementation is available at \url{https://github.com/Enri2077/realtime-manifold-mesh-reconstructor}


\section{RELEVANT RELATED WORKS}
\label{sec:related}
Real-time 3D mapping has been addressed with many different approaches.
The seminal work in \cite{pollefeys_et_al_08} fuses depth maps and it creates an initial 3D rectangle mesh made by two triangles that covers all the depth images, then each triangle is subdivided iteratively where a discontinuity or a non-planarity in the depth map region enclosed in the triangle occurs.
The meshes generated for each depth map are finally registered and the redundant triangles are deleted. This approach does not handle very well occlusions and it requires GPU computing.
Bandino et al. \cite{badino2009stixel} reconstruct an urban scene by  approximating vertical 3D surfaces with adjacent rectangular sticks; as in the previous case, they exploit depth maps computed by means of GPU processing and do not recover a complete model of the scene.

Recently, Vineet et al. ~\cite{vineet2015icra} fuse semantic segmentation and stereo data to obtain a robust estimate of the environment model via Conditional Random Fields (CRF) optimization; their work is close to real-time, but it still requires a significant computational effort.
Similarly, the authors of~\cite{schops20153d} have proposed a real-time reconstruction algorithm which fuses dense maps into a Truncated Distance Signed Function (TDSF) represented by an octree data structure; they estimate the surface only where strong evidence has been collected, therefore the output is non continuous and more prone to localization, and path planning errors, e.g., where the ground is missing. 
%
Following a similar approach Klingensmith et al. \cite{klingensmith2015chisel} fuses dense maps into a TDSF, but they use RGB-D data not available in outdoor scenarios.
In \cite{xu2015real} the authors propose an efficient real-time navigation platform that fuses FPGA stereo disparities into an integer voxel-based maps. 
Even if they build a local map  on a CPU, it is very noisy and prone to localization errors and it has been tested in limited real-world scenarios.

A different class of reconstruction algorithms subdivides the space into tetrahedra through Delaunay Triangulation; this avoids unnecessary memory consumption and results in fast and scalable algorithms.
Online mesh reconstruction of a small object from sparse data that relies on Delaunay triangulation has first been proposed in \cite{pan_et_al09}, however this method recomputes the map at each iteration, it is therefore not suitable for large-scale environments.
The first incremental approach was proposed by Lovi et al.~\cite{lovi_et_al_11}; in their work they create a Delaunay Triangulation incrementally and they update the labels of the  tetrahedra according to the visibility rays from cameras to points. 
The mesh extracted by the proposed method is the sharp boundary between free space and occupied tetrahedra. 
In a latter work, Hoppe et al.~\cite{hoppe2013incremental} have been able to extract the mesh incrementally with a graph cut algorithm, obtaining smoother results but without reaching real-time performance.
Teixeira and Chli \cite{teixeira2016real} build a 2D triangulation of image features, and project it in 3D; they reconstruct a 3D mesh that covers the region of the scene currently captured by the camera.

The previous incremental methods are not able to enforce the manifold property and therefore further mesh refinements, as a simple smoothing or the  one proposed in~\cite{li2015detail}, would be problematic to obtain.
For this reason Litvinov and Lhuiller~\cite{litvinov_lhuillier_13} proposed an algorithm to update incrementally the mesh while keeping the manifold property valid in all the iterations.
The growing procedure, in that algorithm, carves the 3D space sometimes resulting in visual artifacts that affect the quality of the final reconstruction. This issue was faced in~\cite{litvinov_lhiuller14} where the same authors proposed an ad-hoc method to explicitly remove visual artifacts and in~\cite{romanoni15b} where the authors changed the ordering of the carving procedure in order to preemptively remove problematic tetrahedra, i.e., those most likely to result in visual artifacts.

The main issue with all these approaches is that they are not able to reach real-time performance, even if their main goal, beside mapping, is to support a robot while navigating autonomously.
Moreover, they are not able to cope with modifications of the 3D points estimates which is the common situation in SLAM algorithms; a first attempt to overcome the latter issue was proposed in~\cite{romanoni15a} where a simple, but effective, heuristic has been proposed to update the triangulation coherently when moving points already in it. However it is still an approximation and in the long term it may cause drifting of the tetrahedra labeling.

In Table \ref{tab:soa} we summarized the features of the state-of-the-art approaches presented in this Section. Since we are mainly interested in 3D manifold meshes, in the following we focus our comparison against the algorithms proposed in~\cite{litvinov_lhuillier_13} and improved in~\cite{romanoni15b}.
\begin{table}[t]
\caption{State-of-the-art 3D reconstruction algorithms}
\label{tab:soa}
\centering
\setlength{\tabcolsep}{1px}
\begin{tabular}{lcccccccccccc}
\toprule 
& 
\cite{pollefeys_et_al_08} & 
\cite{badino2009stixel} & 
\cite{vineet2015icra} & 
\cite{schops20153d} & 
\cite{klingensmith2015chisel} & 
\cite{xu2015real} & 
\cite{pan_et_al09} & 
\cite{lovi_et_al_11} & 
\cite{hoppe2013incremental} & 
\cite{teixeira2016real} & 
\cite{litvinov_lhuillier_13} &
\cite{romanoni15b} \\
\midrule
Real-time 		& \cmark & \cmark & \cmark & \cmark & \cmark & \cmark & \xmark & \xmark & \xmark & \cmark & \xmark & \xmark \\
CPU-only 		& \xmark & \xmark & \xmark & \cmark & \xmark & \cmark & \cmark & \cmark & \cmark & \cmark & \cmark & \cmark \\
Large-Scale 	& \cmark & \cmark & \cmark & \cmark & \cmark & \xmark & \cmark & \cmark & \cmark & \cmark & \cmark & \cmark \\
Manifold Mesh 	& \xmark & \xmark & \xmark & \xmark & \xmark & \xmark & \xmark & \xmark & \xmark & \xmark & \cmark & \cmark \\
Continuous Mesh & \xmark & \xmark & \cmark & \xmark & \xmark & \cmark & \cmark & \cmark & \cmark & \cmark & \cmark & \cmark 
\end{tabular}
\end{table}

\section{STATE OF THE ART INCREMENTAL MANIFOLD RECONSTRUCTION}
\label{sec:manifold}

In this section we show a brief overview of the manifold reconstruction algorithm first proposed in~\cite{litvinov_lhuillier_13} and improved in~\cite{romanoni15b} and which is at the basis of our CPU-based real-time approach. 
It relies on a volumetric representation of the environment and it processes batches of points, cameras, and visibility rays, which are  estimated incrementally by a feature-based SLAM algorithm, such as PTAM \cite{klein_murray07} or ORB-SLAM \cite{mur2015orb}.

The algorithm builds a 3D Delaunay triangulation of the input SLAM points; the reconstructed surface, $\delta(O)$, partitions the triangulation between the set $O$ of \emph{Outside} tetrahedra, i.e., the manifold subset of the free space (not all free space tetrahedra would be included in this set), and the complementary set $I$ of \emph{Inside} tetrahedra, i.e., the remaining tetrahedra that roughly represent the matter. 
The notation $\delta(O)$ means the boundary of the set $O$.
This manifold is updated as new points and camera positions are estimated. In this process, tetrahedra are associated to a weight $w$ and they are considered as free space if $w>T_{free}$. Notice that in~\cite{litvinov_lhuillier_13} this weight is simply the counter of the rays that intersects the tetrahedron while in~\cite{romanoni15b} the weight is computed according to the inverse cone heuristics.

\begin{figure}[tp]
\centering
\setlength{\tabcolsep}{1px}
 \begin{tabular}{>{\centering}m{0.31\columnwidth} >{\centering}m{0.31\columnwidth} >{\centering\arraybackslash}m{0.31\columnwidth}}
    \includegraphics[width=0.3\columnwidth]{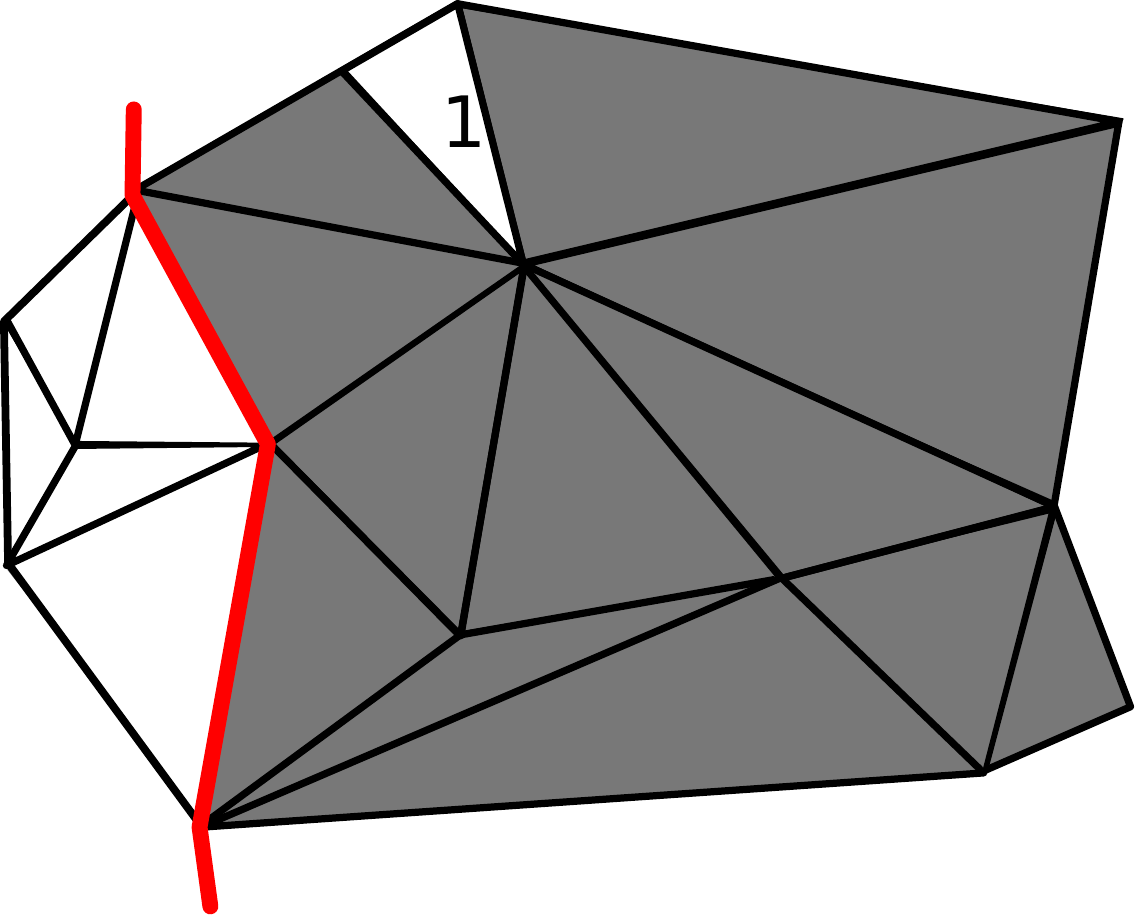}&
  \includegraphics[width=0.3\columnwidth]{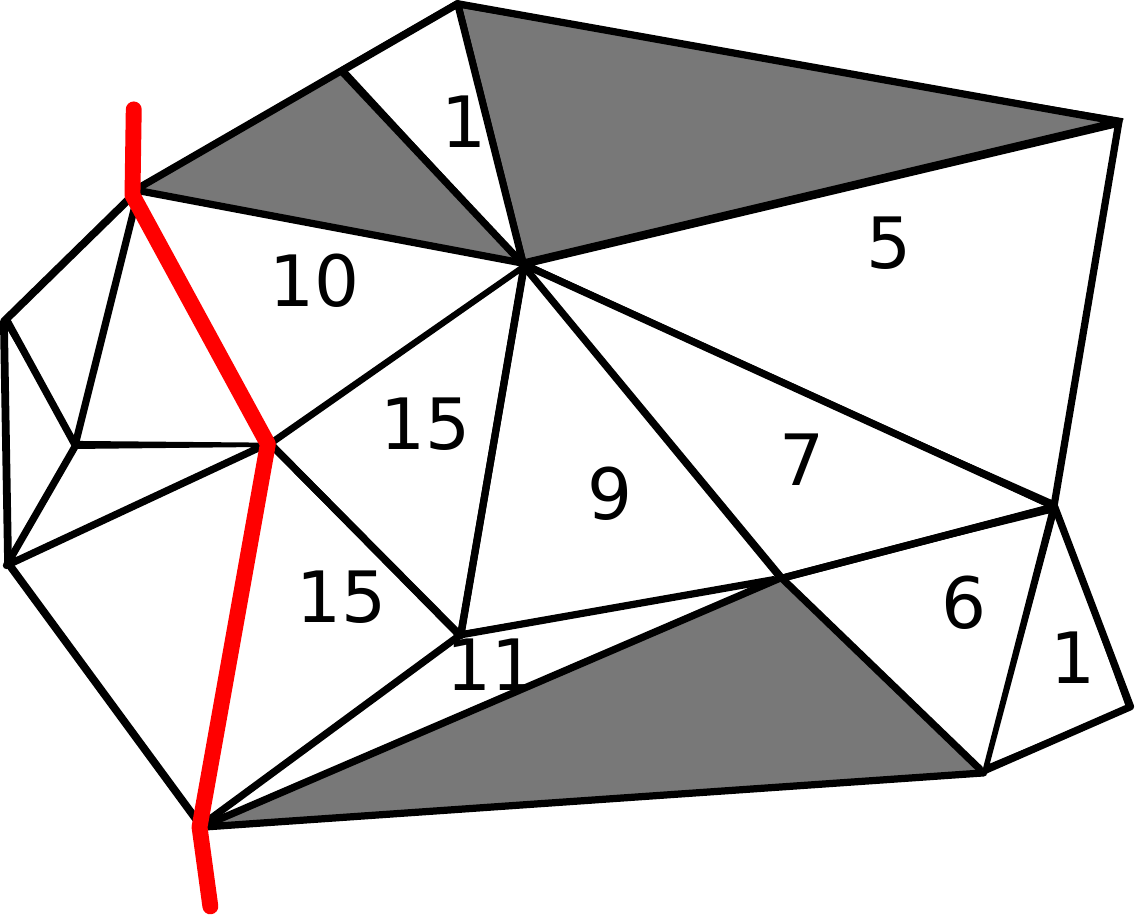}&
  \includegraphics[width=0.3\columnwidth]{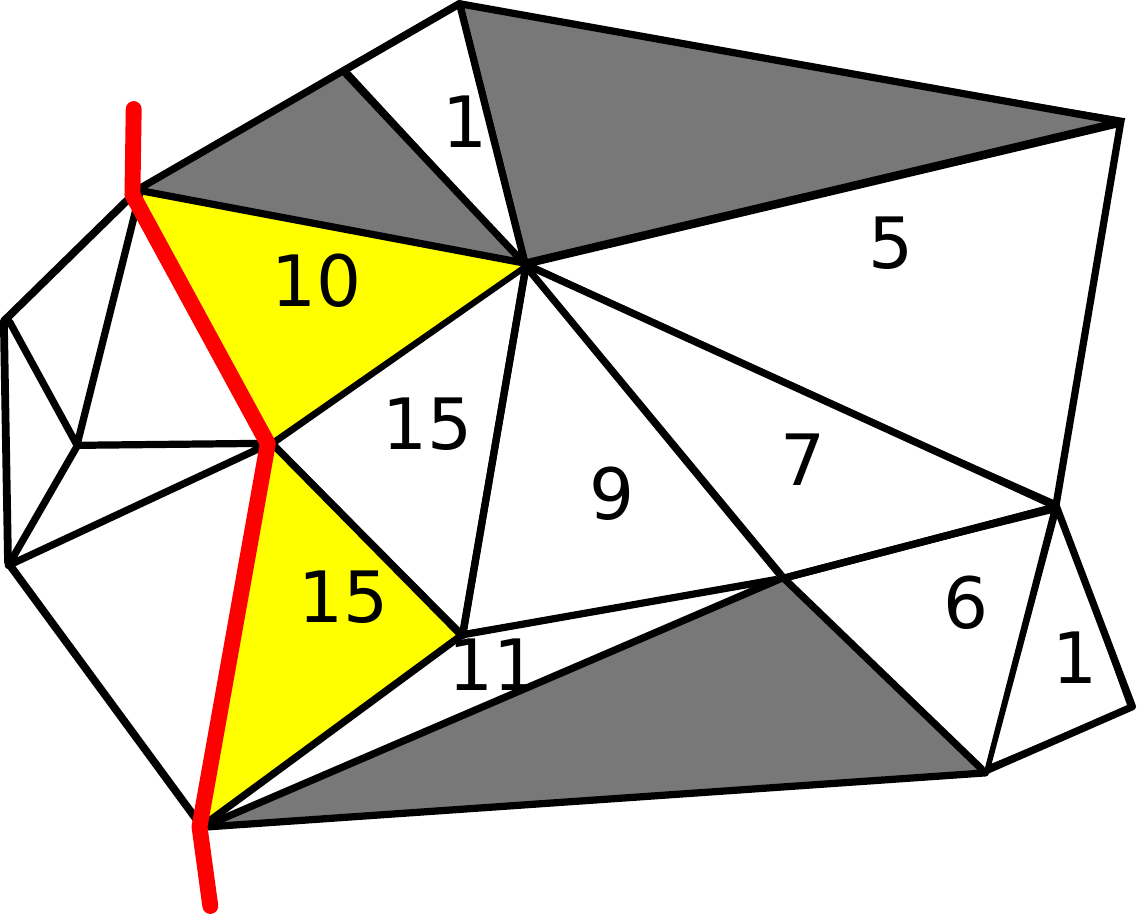}\\
  \includegraphics[width=0.3\columnwidth]{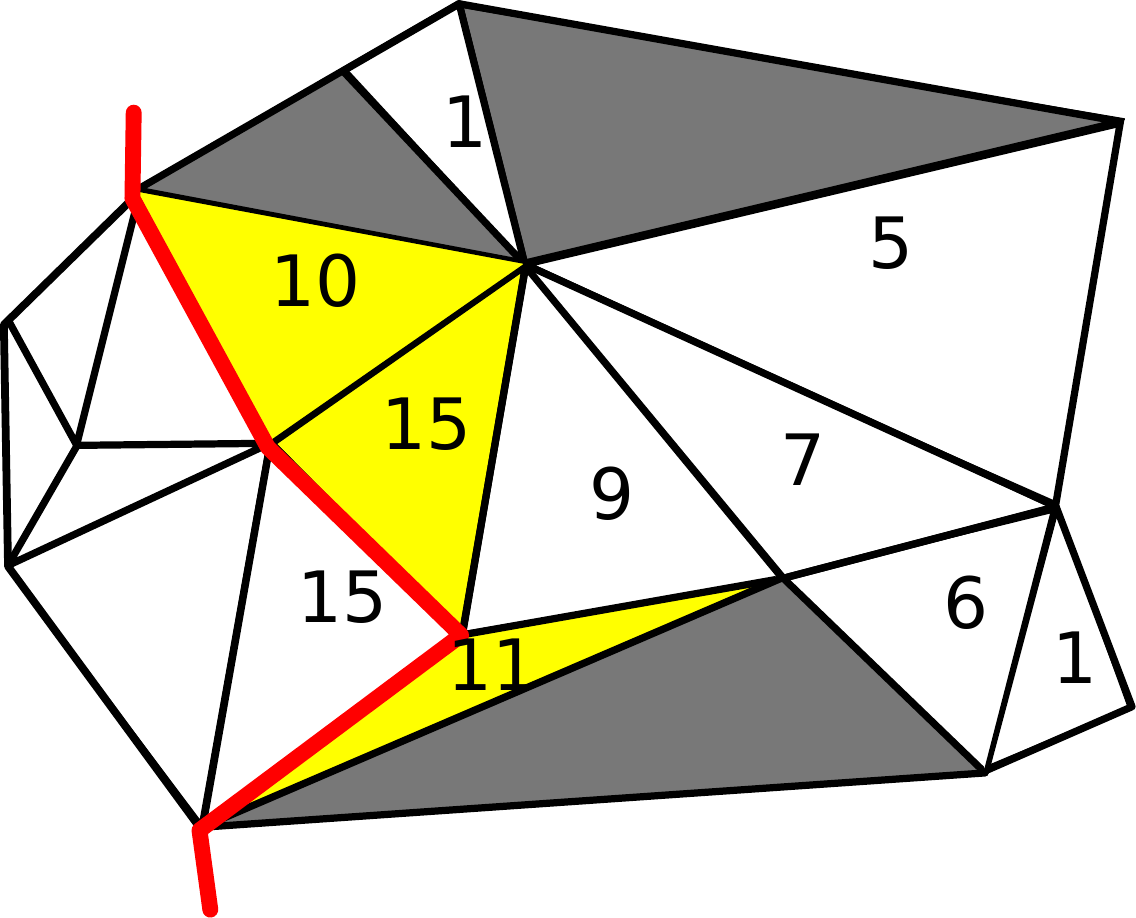}&
  {\Huge...}
  &
  \includegraphics[width=0.3\columnwidth]{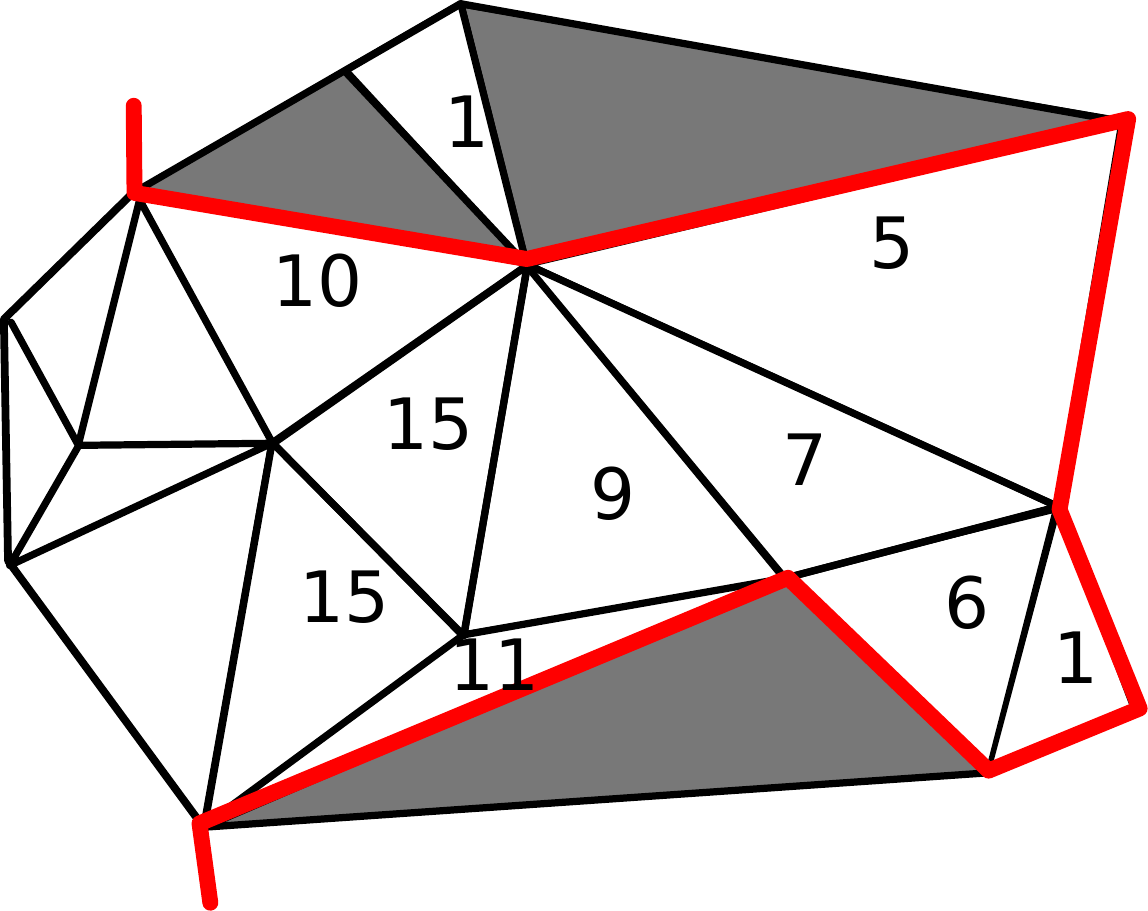}
 \end{tabular}
 \caption{Example of manifold growing. The red line is the manifold, numbers represent the weights, in yellow are the tetrahedra in the queue $Q$, in dark gray the tetrahedra not intersected.}
 \label{fig:growing}
\end{figure}

The first manifold at time $t_{\text{init}}$, named $\delta( O_{t_{\text{init}}})$, is estimated in four steps.
\emph{Point insertion}: it adds all the 3D points estimated up to time $t_{\text{init}}$ and computes their 3D Delaunay triangulation.
\emph{Label initialization}: it initializes all the tetrahedra weight to $0$.
\emph{Ray tracing}: for each camera to point ray it adds a weight $w_1$ to the intersected tetrahedra and a weight $w_2$ to their neighbors, this to preemtively remove most visual artifacts (see \cite{romanoni15b}); the list of free tetrahedra is named $F_{t_{\text{init}}}$.
\emph{Growing}: it initializes a queue $Q$ with the tetrahedron $\Delta_1 \in F_{t_{\text{init}}}$ with the highest weight; then it iterates the following procedure until $Q$ is empty: (a) remove the tetrahedron $\Delta_{\text{curr}}$ with the highest weight from $Q$; (b) add it to $O_{t_{\text{init}}}$ only if the resulting $\delta (O_{t_{\text{init}}} \cap \Delta_{\text{curr}})$  is manifold; (c) add to the queue $Q$ neighboring tetrahedra of $\Delta_{\text{curr}}$ that are not already in the queue or in the $O_{t_{\text{init}}}$ set (see Figure~\ref{fig:growing}). 
Finally, to handle non-zero genus structures, it checks, for each vertex in the boundary, if all the incident tetrahedra $\mathbf{\Delta}_{\text{curr}}$ are free-space and it  tries to remove them; if $\delta (O_{t_{\text{init}}} \cap \mathbf{\Delta}_{\text{curr}})$ is manifold, i.e., all the vertices in $\mathbf{\Delta}_{\text{curr}}$   keeps the manifoldness, then $O_{t_{\text{init}}} = O_{t_{\text{init}}} \cap \mathbf{\Delta}_{\text{curr}}$, otherwise it keeps $O_{t_{\text{init}}}$ unchanged.

\begin{figure}[tp]
\centering
 \begin{tabular}{cc}
  \includegraphics[width=0.4\columnwidth]{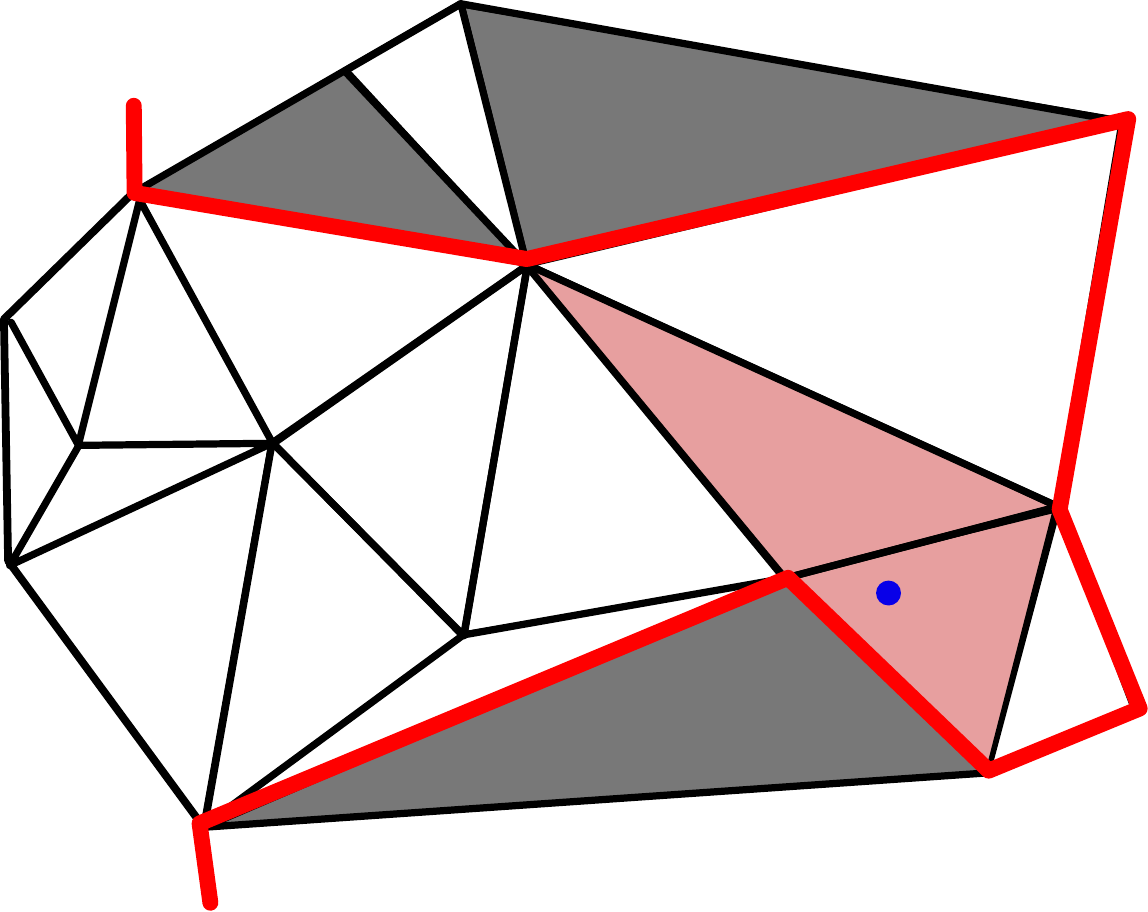}&
  \includegraphics[width=0.4\columnwidth]{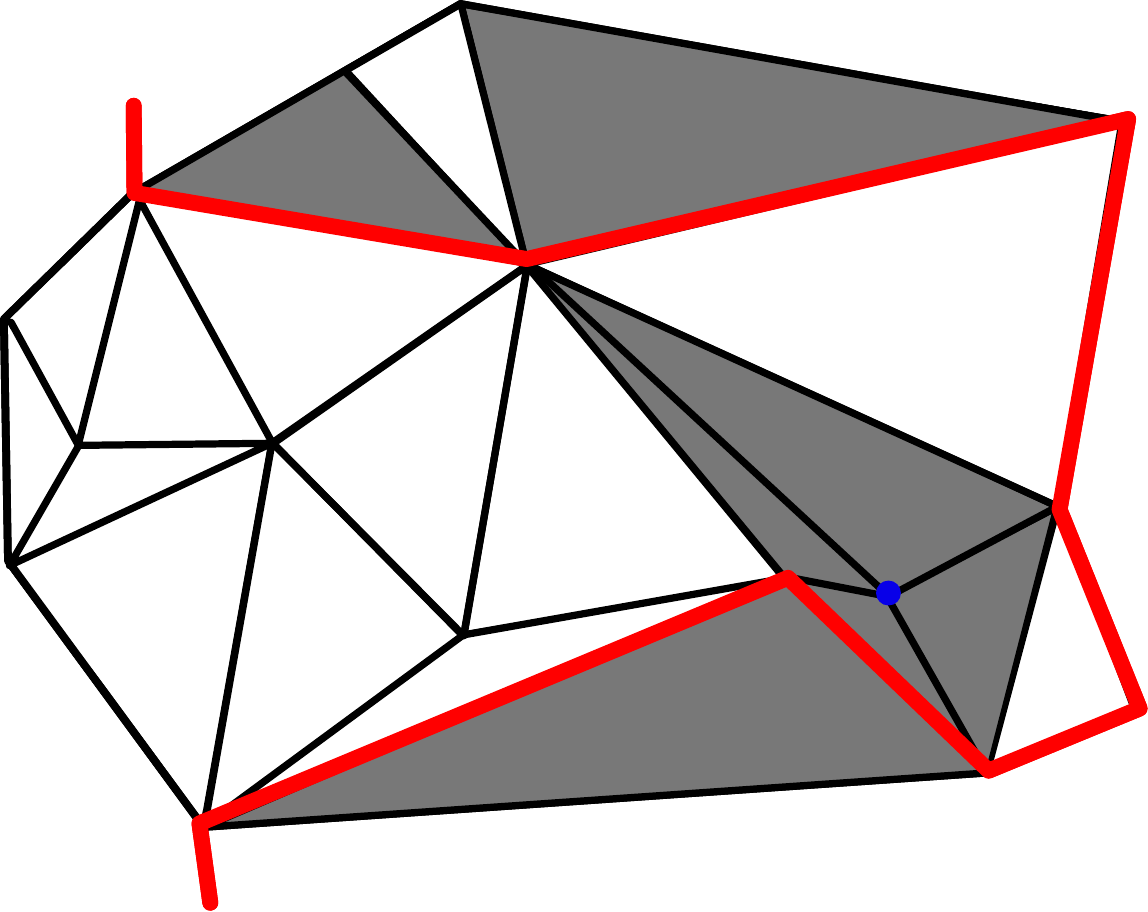}\\
  (a)&(b)
 \end{tabular}
 \caption{Naive point insertion: a new point added to the triangulation triggers a new triangulation of a subset of the tetrahedra; it is not trivial to infer the new tetrahedra status (inside/outside) keeping the manifoldness valid without a new ray tracing.}
 \label{fig:insertion}
\end{figure}

Once the system is initialized, a new set of points $P_{t_k}$ is estimated at time $t_k= t_{\text{init}} + k*T_k$, where $k \in \mathbb{N^+}$ is the keyframe index, and $T_k$ is the period (in our experiments $T_k$ varies and depends on the SLAM algorithm, however it is usually about 4 frames). In this context with the term keyframe we refer to a batch of points and cameras to be inserted in the Delaunay triangulation. 
The insertion of each point $p\in P_{t_k}$ would cause the removal of a set $D_{t_k}$ of the tetrahedra that invalidates the Delaunay property (Figure \ref{fig:insertion}(a)); the surface $\delta (O_{t_k}) = \delta (O_{t_{k-1}} \setminus D_{t_k})$ is not guaranteed to be manifold anymore and any naive update of the triangulation potentially invalidates the manifoldness (Figure \ref{fig:insertion}(b)). 

\begin{figure}[tp]
\centering
 \begin{tabular}{ccc}
  \includegraphics[width=0.3\columnwidth]{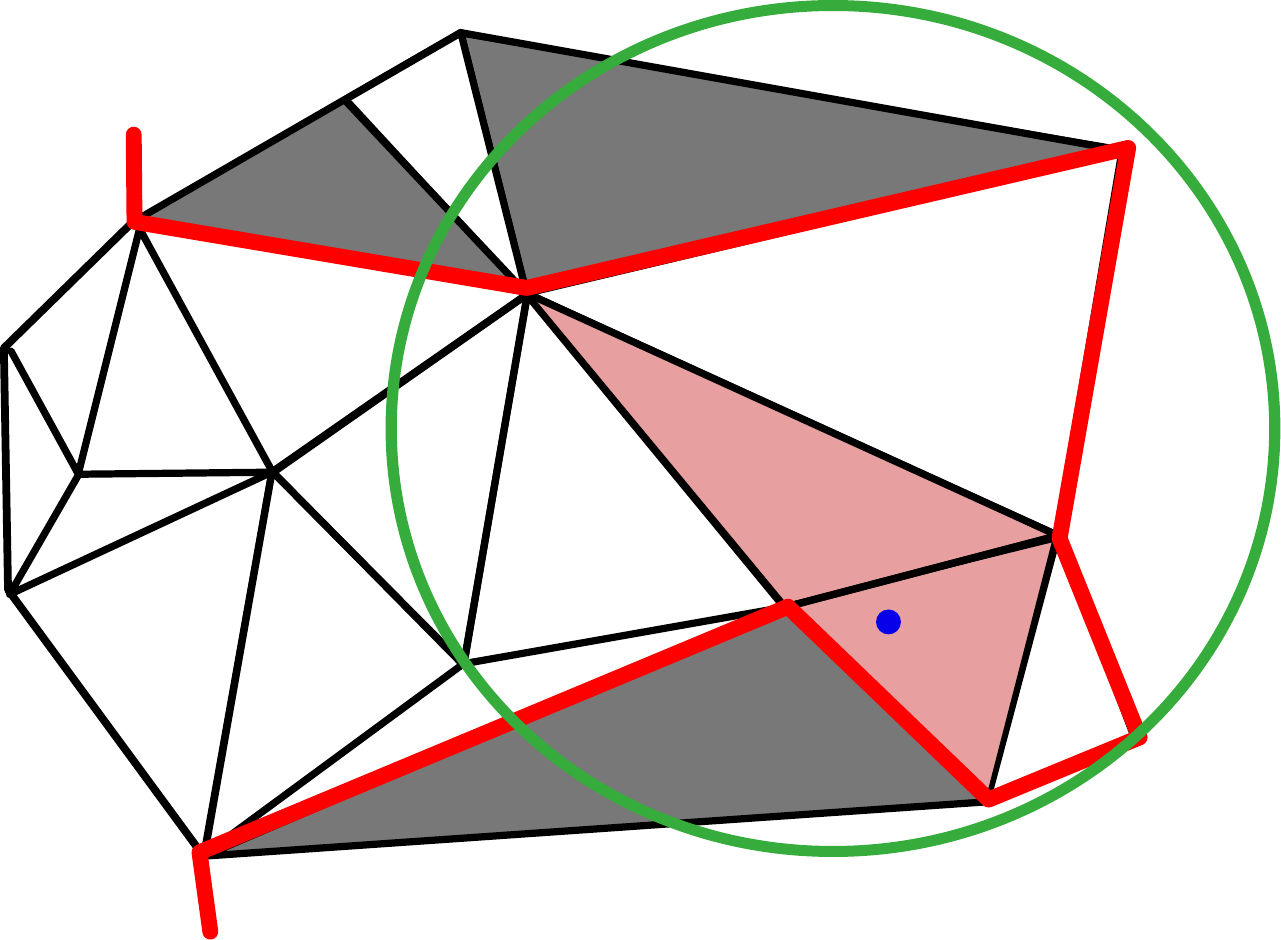}&
  \includegraphics[width=0.3\columnwidth]{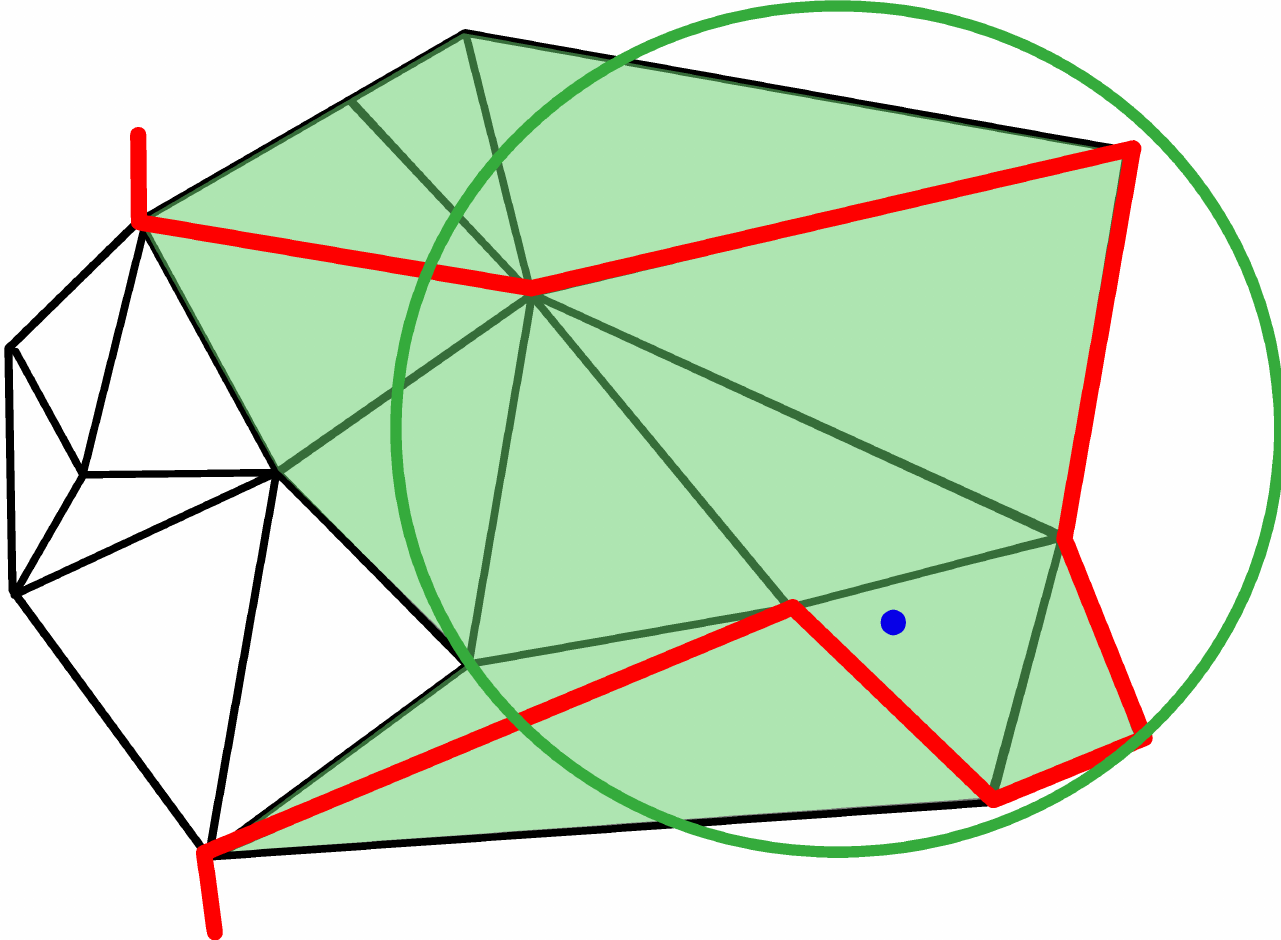}&
  \includegraphics[width=0.3\columnwidth]{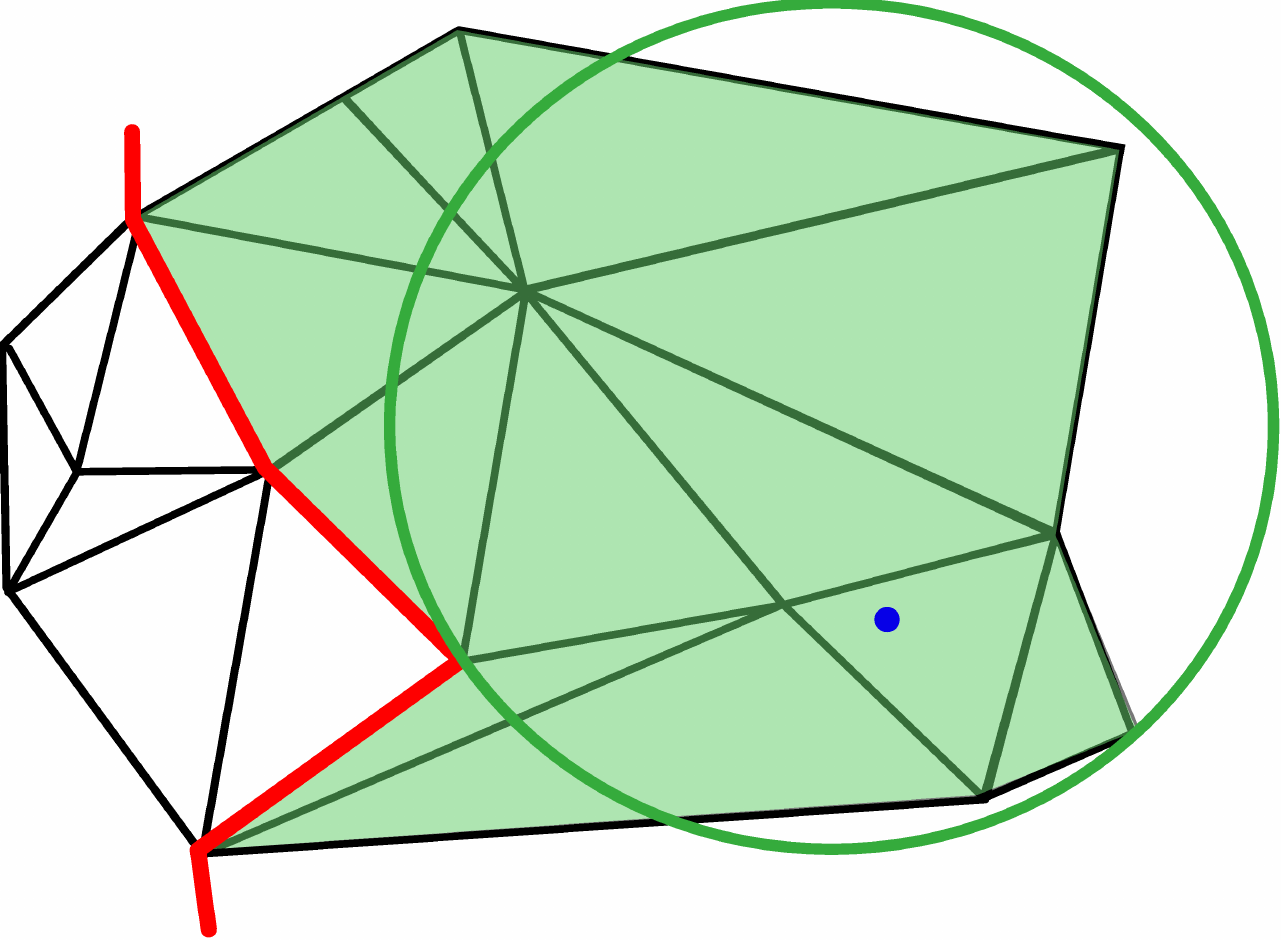}\\
  (a)&(b)&(c)\\
  \includegraphics[width=0.3\columnwidth]{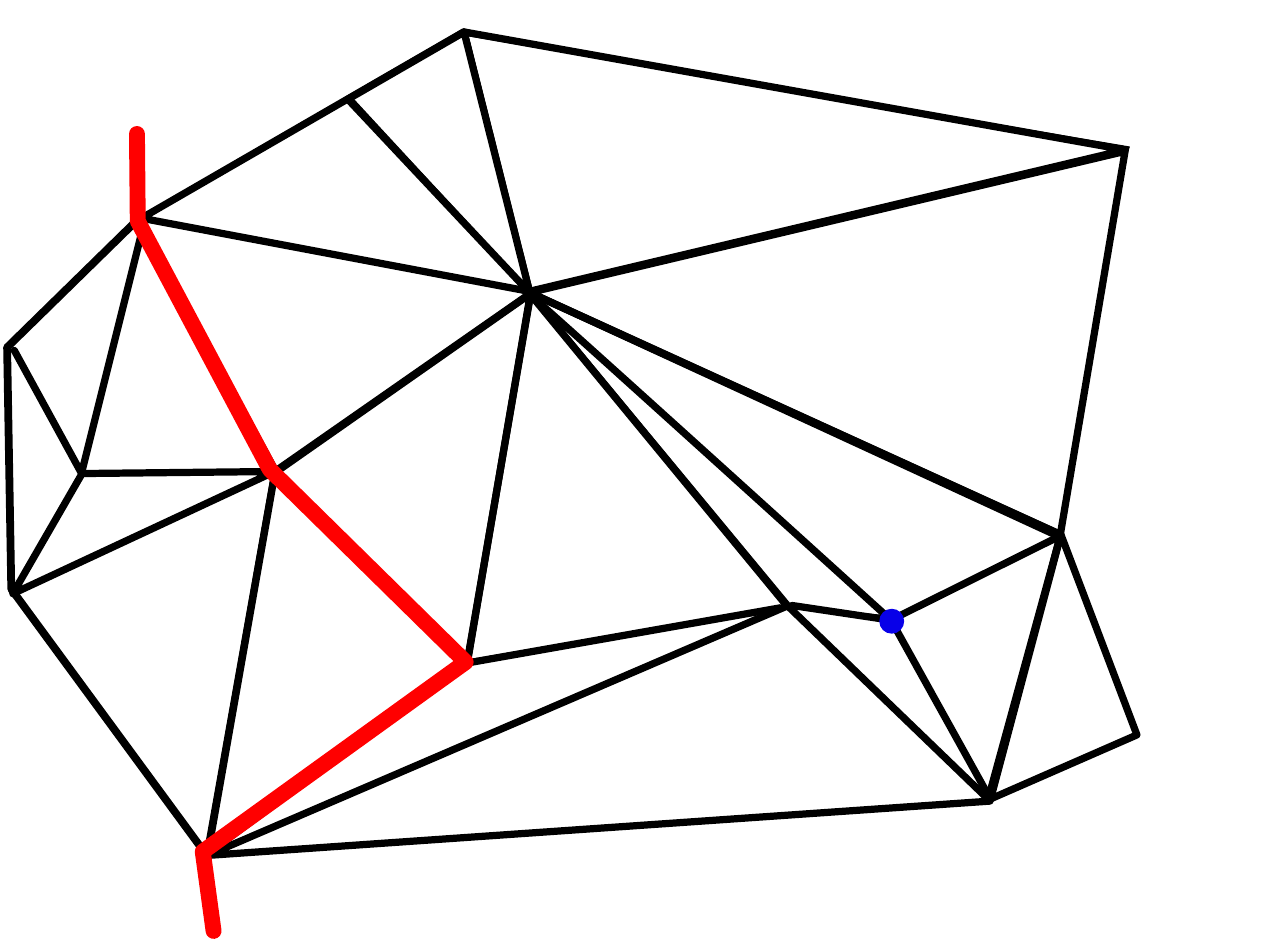}&
  \includegraphics[width=0.3\columnwidth]{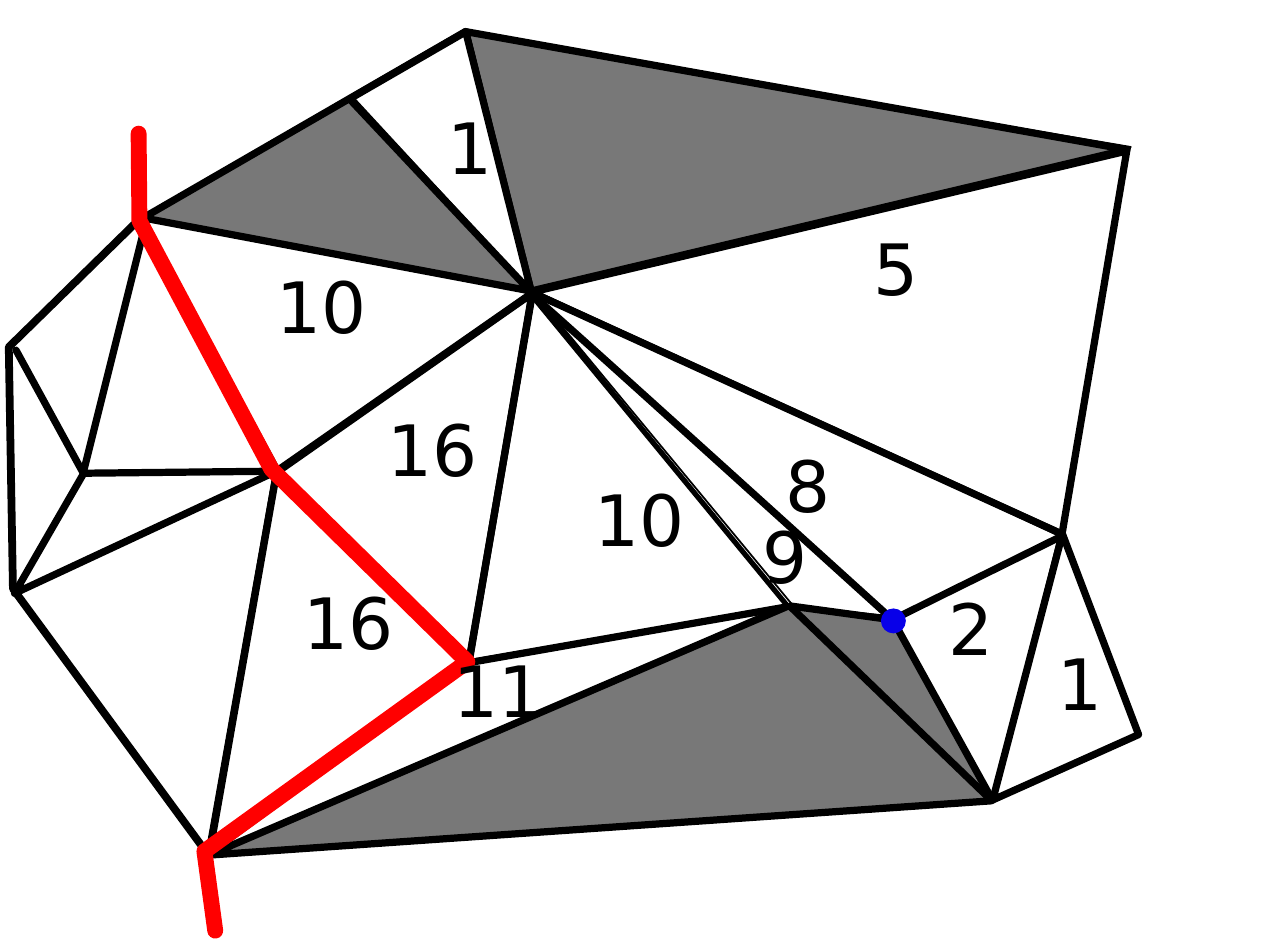}&
  \includegraphics[width=0.3\columnwidth]{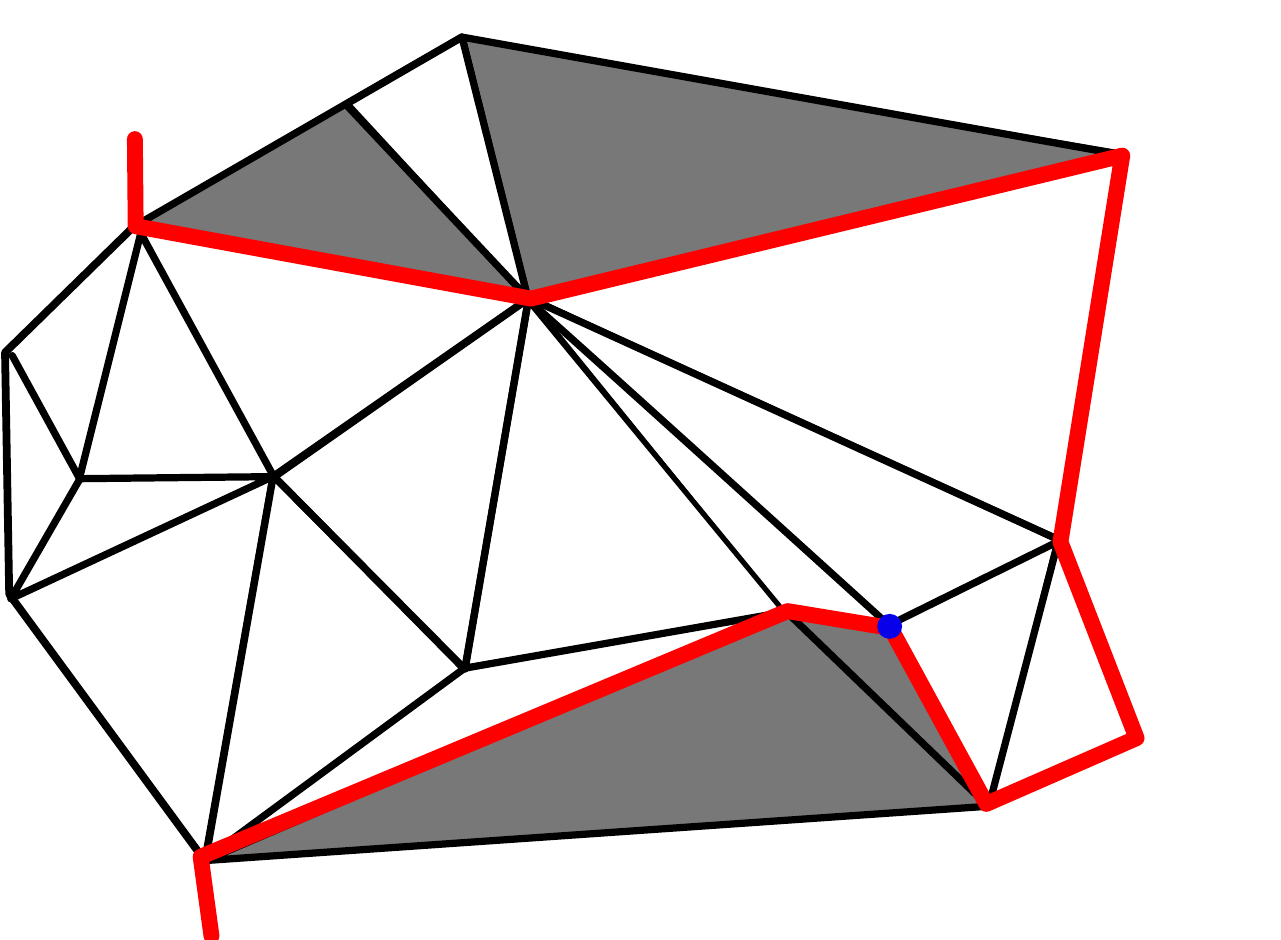}\\
  (d)&(e)&(f)\\
 \end{tabular}
 \caption{Incremental manifold reconstruction steps: (a) a new point is estimated and it would invalidate the red tetrahedra (set $D_{t_k}$); (b) put a subset of the tetrahedra in $E_{t_k}$; (c) shrink the manifold; (d) add the point into the triangulation; (e) perform the ray tracing; (f) grow the manifold.}
 \label{fig:manifoldreconstruction}
\end{figure}

To avoid manifoldness violation, the authors in~\cite{litvinov_lhuillier_13} define a new set of tetrahedra $E_{t_k} \supset D_{t_k}$, named \textit{enclosing set}, (Figure \ref{fig:manifoldreconstruction}(b)) and they apply the so called \emph{Shrinking} procedure first (Figure \ref{fig:manifoldreconstruction}(c)), i.e., the inverse of Growing; this procedure subtracts iteratively from $O_{t_{k-1}}$ the tetrahedra  $\Delta \in E_{t_k}$ keeping the manifoldness valid.
After this process, it is likely that $D_{t_k} \cap O_{t_k} = \emptyset$; however, in the case of $D_{t_k} \cap O_{t_k} \neq \emptyset$ the point $p$ is not added to the triangulation and it is discarded.
In case of $D_{t_k} \cap O_{t_k} = \emptyset$ the point is added into the triangulation with the \emph{Point Addition} step (Figure \ref{fig:manifoldreconstruction}(d)), and once all points in $P_{t_k}$ have been added (or dropped), the algorithm applies Ray tracing (Figure \ref{fig:manifoldreconstruction}(e)), and finally the Growing step (Figure \ref{fig:manifoldreconstruction}(f)) similarly to what is described in the initialization procedure.

To keep the reconstruction scalable, both \cite{litvinov_lhuillier_13} and \cite{romanoni15b} create a grid of fictitious 3D points, named Steiner points, which spans the entire space the map is supposed to occupy. 
Steiner points partition the space using big cubes and the cube diagonal becomes the maximum diameter of the circumsphere used to verify the Delaunay property of the triangulation; during the estimation of the enclosing set, the cube diagonal defines a bounded set of tetrahedra affected by the point insertion.

\section{REAL-TIME MESH RECONSTRUCTION}
\label{sec:overview}
In this section we illustrate the main contribution of this paper, i.e., we describe how we build upon the previous method in order to achieve real-time performances and to extend it to moving point management.
We redesigned some steps of the reconstruction pipeline proposed in \cite{litvinov_lhuillier_13} and \cite{romanoni15b}, mostly by leveraging on the use of Steiner points, which are arbitrary 3D points added to the 3D Delaunay triangulation to bound the maximum dimension of the tetrahedra.
Even if some of the methods we propose are designed to optimize \cite{litvinov_lhuillier_13} and \cite{romanoni15b}, some of the concepts can be easily adapted to improve the efficiency of other Delaunay-based reconstruction algorithms.

\subsection{Incremental Steiner points insertion}
The original method initializes the algorithm with a fixed grid of Steiner points, instead, we estimate the Steiner points grid iteratively keeping the manifold property valid and while new parts of the scene are discovered. 

Let us define $l_{\text{Steiner}} = 10m$ as the distance among the Steiner points in the grid; we keep track of the boundaries of the current grid by saving $(x_{\text{bound}}^{\text{min}}, y_{\text{bound}}^{\text{min}}, z_{\text{bound}}^{\text{min}})$ and $(x_{\text{bound}}^{\text{max}}, y_{\text{bound}}^{\text{max}}, z_{\text{bound}}^{\text{max}})$.
We bootstrap with an initial grid around the first camera pose $c_0$, then, at each iteration $i$, if the points visible by camera $c_i$ span outside the existing grid, we expand it.
More formally, before adding a point in position $p = (x, y, z)$, we check if $x_{\text{bound}}^{\text{min}} < x < x_{\text{bound}}^{\text{max}}$,  $y_{\text{bound}}^{\text{min}} < y < y_{\text{bound}}^{\text{max}}$ and  $z_{\text{bound}}^{\text{min}} < z < z_{\text{bound}}^{\text{max}}$; in case all these conditions are verified, the point can be added without extending the grid; otherwise we extend the grid along the appropriate semi-axis, e.g., if $x < x_{\text{bound}}^{\text{max}}$, the grid is extended on the positive $x$ semiaxis by inserting a layer of
Steiner points with coordinate $x = x_{\text{bound}}^{\text{max}} + l_{\text{Steiner}}$.


\subsection{Shrink what you need}
The efficiency of the shrinking procedure is strictly related to the size of the enclosing set $E_{i}$.
Indeed the set $E_{i}$ contains the tetrahedra that the algorithm need to shrink to keep the manifold property valid before inserting a set of new points.
In the original method of ~\cite{litvinov_lhuillier_13} and~\cite{romanoni15b} the authors fix $r_{\text{max}}$ as the maximum accepted distance  between a camera and point, and they define $E_{c}$ with respect to a camera $c$, as the tetrahedra contained in the sphere centered in $c$ with radius $r_{\text{max}} + sqrt(3)*l_{\text{Steiner}}$

The drawback of this approach is that the enclosing volume is needlessly large, in fact, except for the case of an omnidirectional camera or an omnidirectional laser sensor, the point cloud visible from a camera is not evenly spread in a sphere with radius $r_{\text{max}}$, but it is mostly clustered on the actual surface of the scene perceived by the sensor. 
Therefore unnecessarily large number of tetrahedra requires to be shrunk before adding new points, even if they occupy just a small part of the sphere. Shrinking huge regions  of the space can produce visual artifacts.

\begin{figure}[t]
\centering
  {\def\svgwidth{0.45\textwidth}
  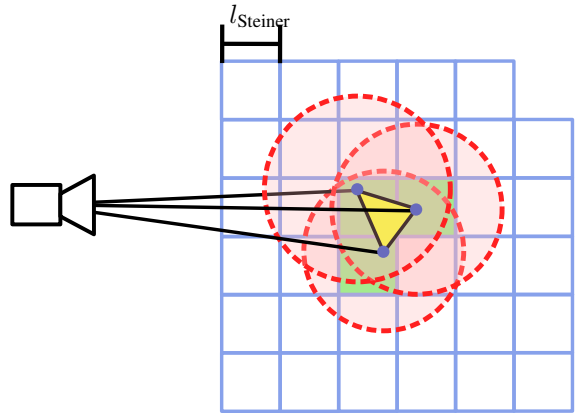}
\caption{Efficient enclosing proposed in this paper}
\label{fig:encl}
\end{figure}

Instead, we propose to choose a smaller enclosing volume, in particular, we define $E_{i}$ as the set of  tetrahedra exactly in the neighborhood of the 3D points that have to be added to the triangulation, regardless of the associated cameras.
Given a set $P$ of new points that have to be inserted, the ideal enclosing volume is the convex hull of $P$, i.e., the yellow region in Figure~\ref{fig:encl}, expanded by a radius of $\sqrt[]{3}\cdot l_{\text{Steiner}}$, i.e., the red region in Figure~\ref{fig:encl}.

We propose to approximate this volume by exploiting the Steiner grid.
Let $p$ be a point contained in $cell_{(i,j,k)}$, where $(i,j,k)$ are the indexes identifying the cell inside the Steiner grid. 
Since a cell diagonal is at most $\sqrt[]{3}\cdot l_{\text{Steiner}}$, the corresponding sphere would span over at most two cells in each direction. 
The enclosing set $E_{i}^p$ induced by $p$ is then obtained by selecting all the tetrahedra that have some vertices inside $cell_{(a,b,c)}$, where $i - 2 \leq a \leq i + 2 \land j - 2 \leq b \leq j + 2 \land k - 2 \leq c \leq k + 2 $.
The final enclosing $E_{i}$ set for the $i$-th camera  is $E_{i} = \bigcup_{p\in P_i} E_{i}^p$.

\subsection{Boundary spatial hashing}
In the original method both the shrinking and growing procedure iterate over the boundary of the manifold mesh in order to subtract or add new tetrahedra to the manifold set $O$.
The lookup procedure that checks if a free space tetrahedra can be carved without invalidating the manifold property, and keeps the boundary ordered, does not scale properly since it has to consider the entire boundary of the mesh.

To improve on this lookup procedure, we propose to store the boundary tetrahedra into a spatial hash representation based on the Steiner grid cells.
We create a hash function $B$ such that $B_(i,j,k)$ is a vector defined for a Steiner cell $cell(i, j, k)$; $B_(i,j,k)$ contains all the pointrs to the boundary tetrahedra that intersects $cell(i, j, k)$.
Thanks to the hash function we can quickly and directly retrieve the tetrahedra needed to initialize the shrinking and growing procedures.

\subsection{Next Tetrahedron Caching}
\begin{figure}[t]
\centering
  {\def\svgwidth{0.48\textwidth}
  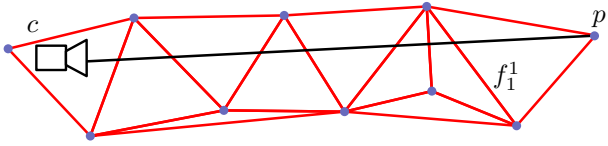}
\caption{Ray tracing from camera $c$ to point $p$}
\label{fig:raytracing}
\end{figure}

The ray tracing described in Section~\ref{sec:manifold} walks along the triangulation  to update the visibility information encoded in the tetrahedra weights.
It bootstraps from the ending point (e.g., $P$ in Figure~\ref{fig:raytracing}) and checks which facets $f_1^1$ of the set of incident tetrahedra intersects the ray; as a consequence, the tetrahedron containing the intersecting facet $f_1^1$ is selected and its visibility information updated; then ray tracing moves to the tetrahedron adjacent to facet $f_1^1$ and iterates the process until the selected tetrahedron contains the camera (e.g., $c$ in Figure~\ref{fig:raytracing}).
Depending on the number of new points inserted per frame, this step may require a big computational effort.

In a robotics scenario, especially when a surveying vehicle explores the environment, the camera-to-point viewing rays have often the same direction, e.g., we keep perceiving a point in the front of the vehicle from the same direction while the camera moves forward. By relying on this assumption, we propose a method to speed up the identification of the next traversed tetrahedron on the walk by Next Tetrahedron Caching.
While walking on a ray, for each traversed tetrahedron $\Delta_i$ we store the index of the next tetrahedron $i_{next}$; then, when we trace any other ray, if in the current tetrahedron the index $i_{next}$ has been saved, the cell $i_{next}$ is checked first. 
Such process likely chooses the right next tetrahedra first, and avoids to check the intersection between the ray and the the other facets if not needed; moreover, by simply prioritizing the tetrahedra checks this caching mechanism is guaranteed to find the same solution of the original ray tracing procedure.

\subsection{Ray tracing scheduler}
The authors in \cite{litvinov_lhuillier_13} and \cite{romanoni15b} do not manage moving points, however, as new images are processed, the point positions estimates are updated by the SLAM algorithm.
Only in~\cite{romanoni15a} a simple heuristic to manage moving points has been proposed, but it approximates the visibility updates induced by the moving points.

To handle moving points in an exact way we use three procedures, analogously to \cite{lovi_et_al_11}:
\textit{Ray tracing} to increment the weights according to the Inverse Cone Heuristic proposed in~\cite{romanoni15b};
\textit{Ray retracing}, which is a particular ray tracing that affects only the new tetrahedra added after point insertion or removal; \textit{Ray untracing}, which decrements the weights and removes the stored ray.
To avoid redundant computations, especially when points positions change, we propose to carefully schedule the ray tracing operations. 
When we add a point $p$, we schedule the tracing procedure for all the rays ending in $p$ and we schedule the ray retracing  of each ray intersecting the conflicting tetrahedra, indeed they have been removed and new tetrahedra have been added, therefore they need to be updated.
When we remove a point $p$, we schedule ray retracing for each ray intersecting the tetrahedra conflicting with $p$, the point is removed, and the ray untracing procedure is scheduled for each ray that connects a camera to the removed point.
When we move a point, we first remove it from the old position and we add it back to the new position.

After we processed  a new frame we collect a list $L_i$ of tracing untracing and retracing procedure for each ray $r_i$. 
To avoid redundancy, $L_i$ contains at most one occurrence of each procedure.
In general we execute the procedures in $L_i$ in this order: ray untracing, ray tracing and ray retracing, subject to the following rules. 
If $L_I$  contains both ray tracing and ray retracing, only the ray tracing is executed, since retracing is equivalent, but it only affects a subset of the tetrahedra.
If $L_I$ contains ray untracing and ray retracing, without ray tracing, the $r_i$ is only untraced. 
This  happens when $r_i$ intersected at least one tetrahedron destroyed by point  insertion or removal, and it is then removed.

During the ray tracing or ray retracing procedures the path of the traversed cells is stored for each ray, avoiding recomputing the path from the camera to the point multiple times.
This reduces significantly the effort needed to move and remove the points, indeed for each of them all the corresponding rays must be untraced and then traced in the new position.

\section{EXPERIMENTAL RESULTS}
\label{sec:experimental}

We tested the proposed real-time incremental manifold mesh reconstruction algorithm on the KITTI dataset~\cite{kitti}; in particular we used four stereo sequences of the visual odometry dataset: 00, 01, 02, and 05, we discarded the 03 and 04 since they are relatively short. 
We executed the experiments on a Intel(R) Core(TM) i7-4770S at 3.10GHz with 8GB of DDR3 RAM with the reconstruction algorithm running on a single core. 

We estimated the 3D points, camera poses and visibility rays through ORB-SLAM~\cite{mur2015orb} which runs on the same machine, using three threads, in real-time. 
Using the ORB-SLAM output we have compared the proposed approach against the two state of the art incremental manifold mesh reconstruction algorithms, i.e.,~\cite{litvinov_lhuillier_13} and~\cite{romanoni15b}. 
In our experiments we discarded the handle and artifacts removal algorithm from~\cite{litvinov_lhuillier_13} since it takes a big computational effort, the number of handles created in these sequences is limited and the smoothing operator mitigates the influence of the artifacts. 
However some of the artifacts remains especially because of the sparsity of the point cloud estimated with ORB-SLAM. 
Since the compared algorithms do not manage loop closures, we considered 1000, 1101, 2000 and 1300 frames, respectively from the sequences 00, 01, 02, 05, which do not have any loop closure.

\begin{figure*}[tp]
\centering
 \begin{tabular}{cccc}
  \includegraphics[width=0.44\columnwidth]{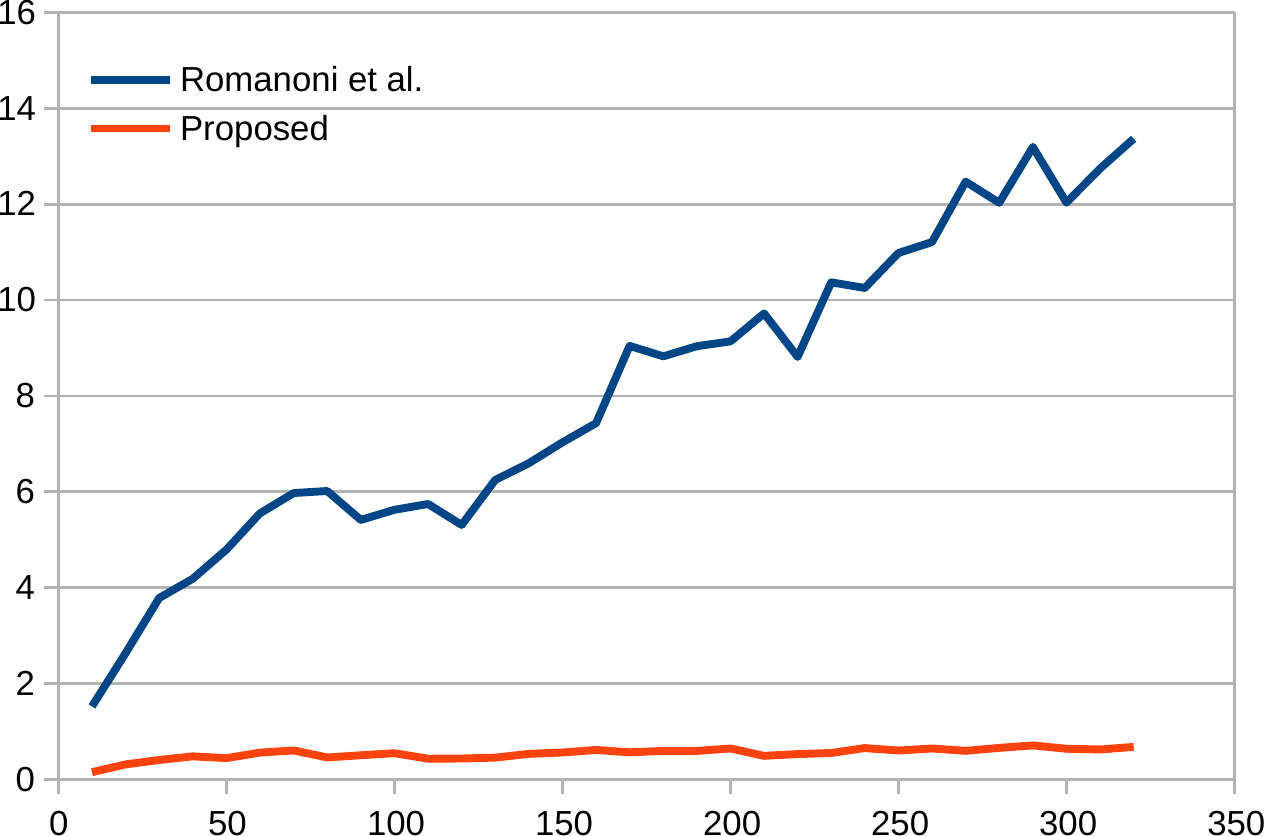}&
  \includegraphics[width=0.44\columnwidth]{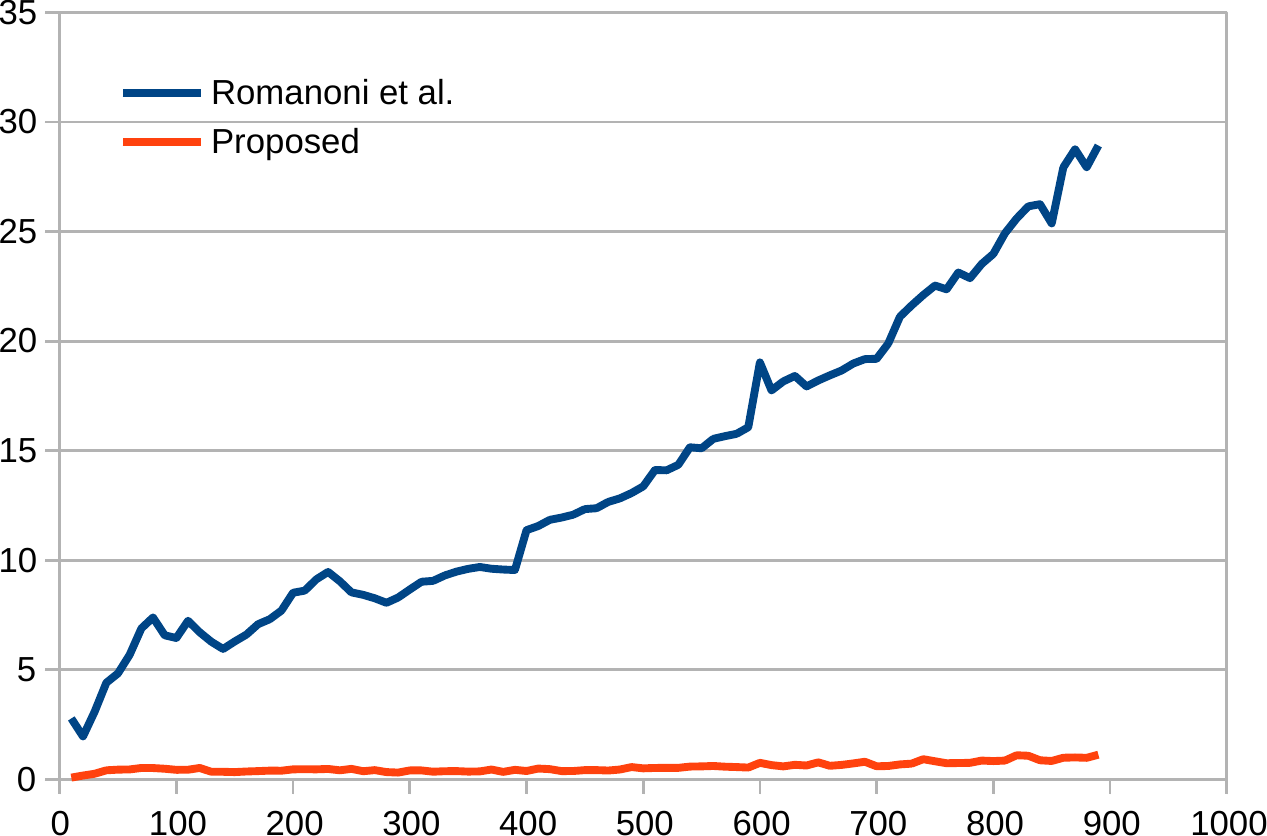}&
  \includegraphics[width=0.44\columnwidth]{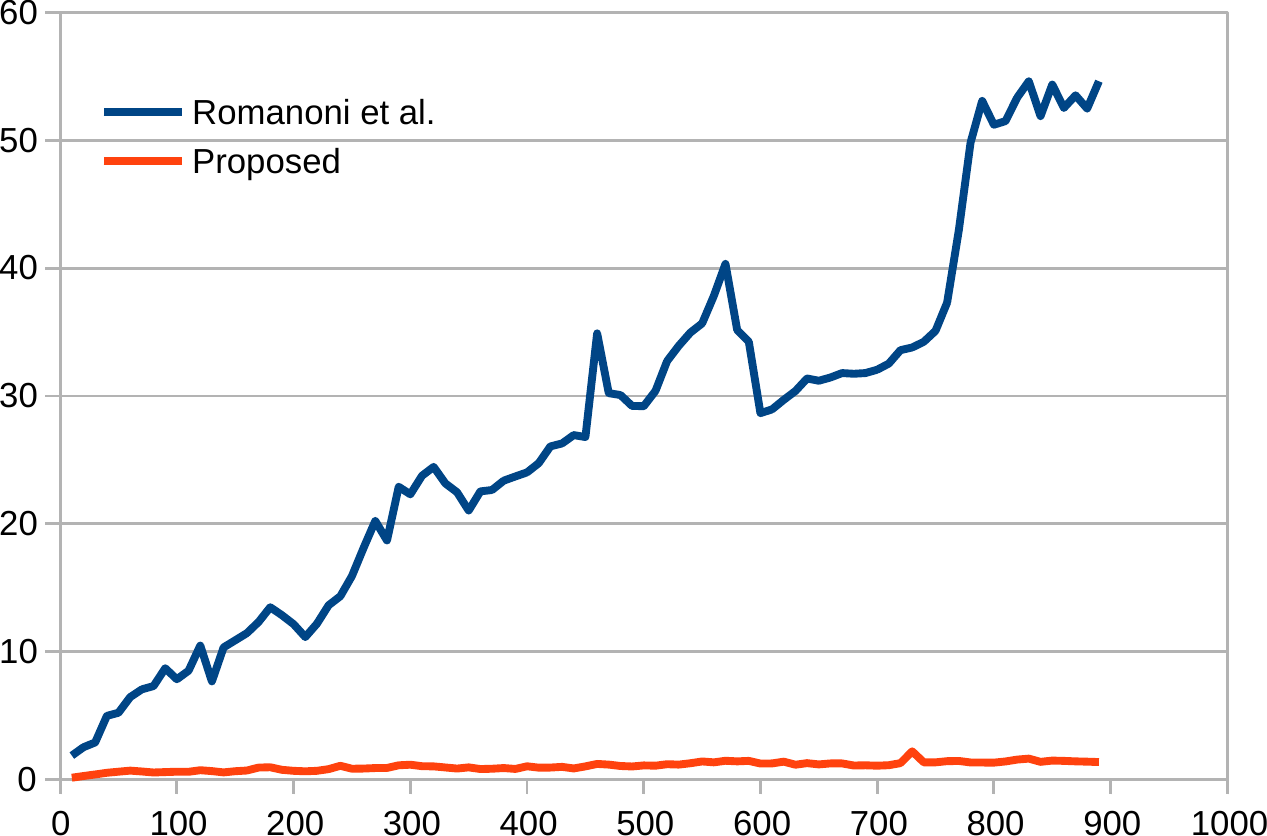}&
  \includegraphics[width=0.44\columnwidth]{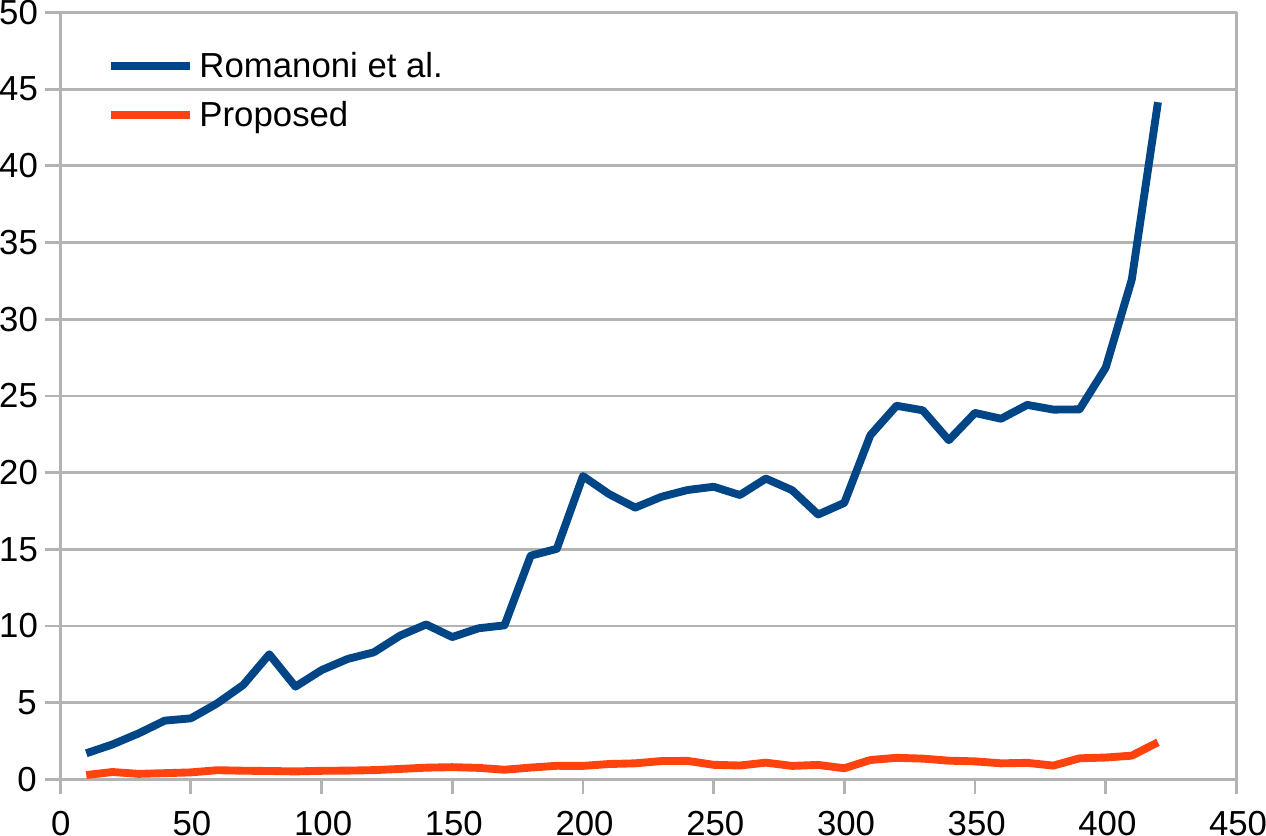}\\
  Sequence 00&
  Sequence 01&
  Sequence 02&
  Sequence 05\\
 \end{tabular}
 \caption{Comparisons of run times between our approach and~\cite{romanoni15b} (x-axis number of keyframe, y-axis: timings in seconds).}
 \label{fig:timingsCompare}
\end{figure*}

The proposed algorithm has been able to reconstruct the test sequences in real-time, improving substantially the computational effort with respect to its competitors.
As in most mapping algorithms that decouple tracking and mapping \cite{klein_murray07,schops20153d,schops2017large}, we build the map every keyframe, i.e., batch of points and cameras, while the ORB-SLAM algorithm tracks the camera position. 
In our case, map updates happen at 3-4 Hz while the tracking threads run at the same frame-frequency of the camera (10Hz).
Comparing the different steps of the reconstruction in Table~\ref{tab:res}, we improved each of them thanks to the novel algorithms described in Section~\ref{sec:overview}. In the last column of Table~\ref{tab:res} we report also the frequency of our algorithm with respect to the total number of frames; as it it can be noticed, the proposed mapping approach is able to process the frames at a frequency which is 4 times higher than the actual data rate.
All techniques used in the novel algorithm have contributed to a substantial improvements, we highlight just that the first tetrahedron proposed by the Next Tetrahedron Caching was actually the correct one the 88\% of the cases.

To visualize the improvement of the proposed method, in Figure~\ref{fig:timingsCompare} and Figure~\ref{fig:timings}, we report the time spent for each keyframe extracted by ORB-SLAM. The run time of~\cite{romanoni15b} grows linearly, especially because of the growing process which execution time increases as the dimension of the map increases. Instead, in our algorithm the boundary spatial hashing bounds significantly the computational cost of the growing procedure and the overall computational time results almost constant. 

\begin{figure}[t]
\centering
 \begin{tabular}{cccc}
  \includegraphics[width=0.45\columnwidth]{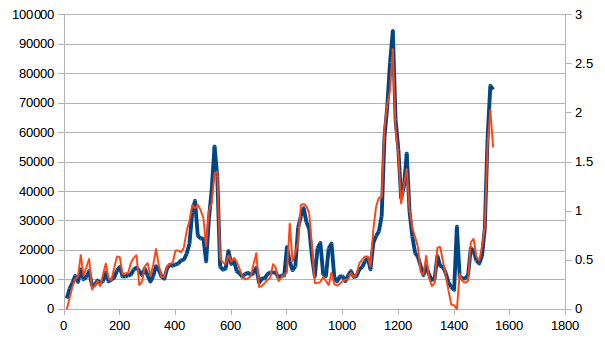}&
  \includegraphics[width=0.45\columnwidth]{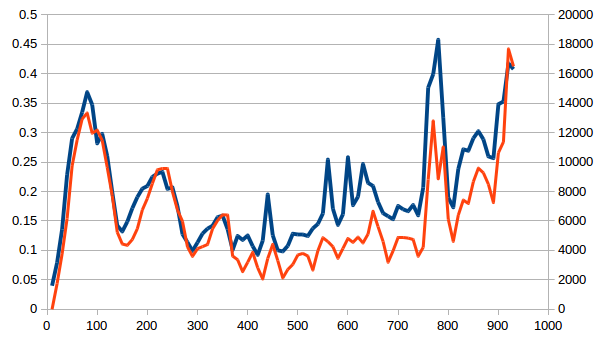}\\
  sequence 00&
  sequence 01\\
  \includegraphics[width=0.45\columnwidth]{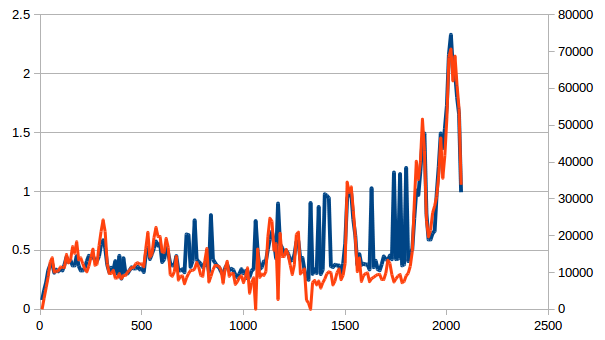}&
  \includegraphics[width=0.45\columnwidth]{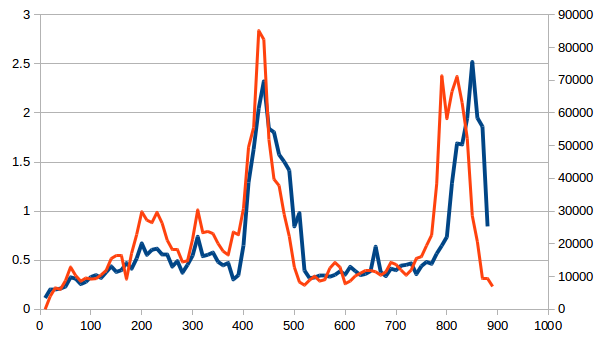}\\
  sequence 02&
  sequence 05\\
 \end{tabular}
 \caption{Run time (blue) vs number of enclosing sets (orange). 
 The graphs shows the high correlation. (x-axis number of keyframe, y-axis (left): timings in seconds, y-axis(right): number of tetrahedra)}
 \label{fig:timingsEncl}
\end{figure}

\begin{figure}[t]
\centering
\includegraphics[width=0.8\columnwidth]{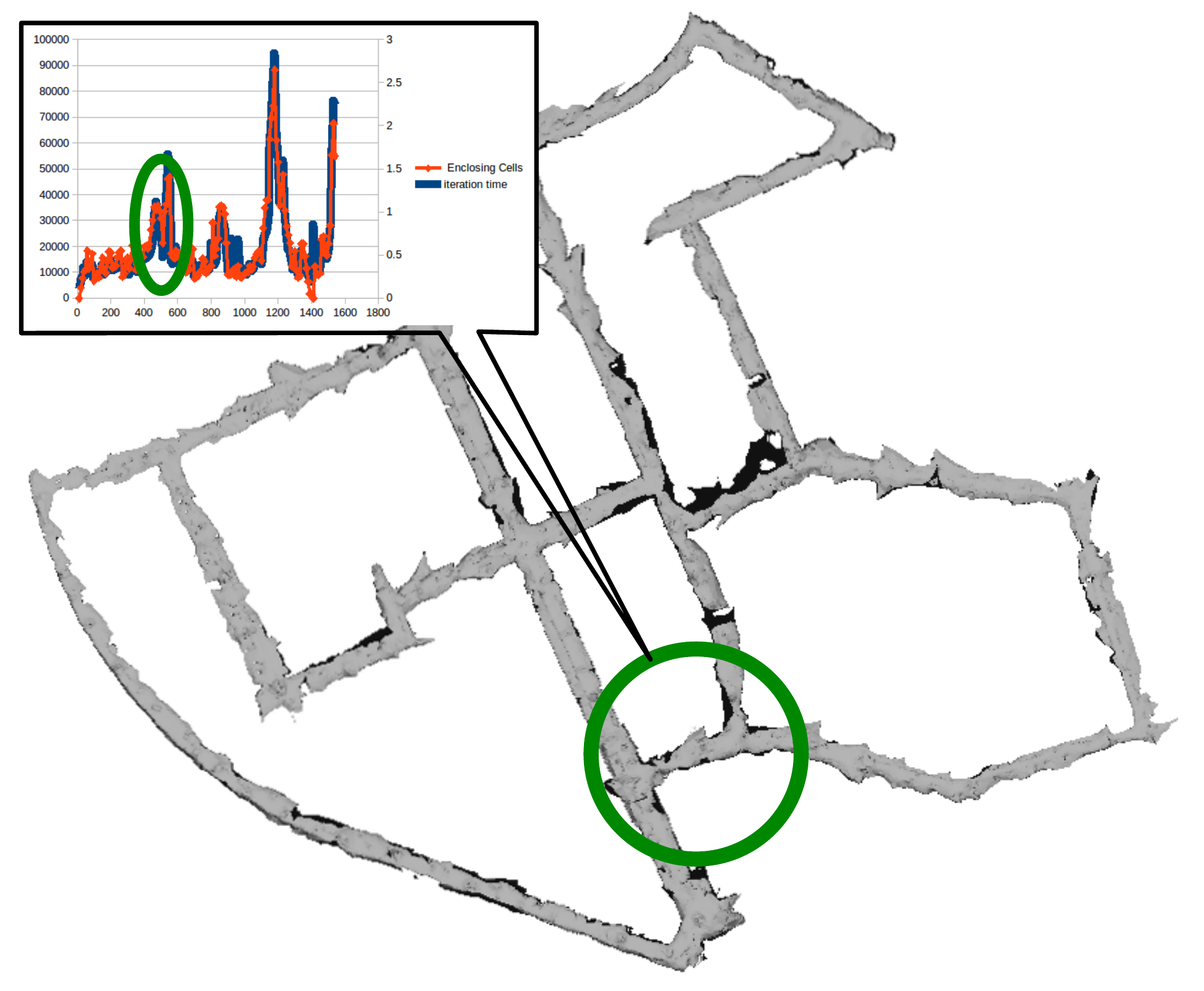}
\caption{Top view (without ceiling) of the reconstruction of the 00 sequence of the KITTI dataset. The plot shows that in the correspondence of genus changes the number of enclosing sets increases, therefore the total timing increases}
\label{fig:kitti00}
\end{figure}

We also tested the proposed approach on the KITTI full sequences, i.e., including parts where a loop closure happens.
In this case we cannot handle explicitly the loop closures and when they happen, small mesh misalignments occur; however, this experiment was designed to proof that the proposed method handles non-zero genus surfaces, and that it scales with the whole sequence.
In Fig. \ref{fig:kitti00} we illustrate for instance the reconstruction of the complete sequence 00, and we plot the  run time in Fig.~\ref{fig:timings-complete-sequences}.
In Fig. \ref{fig:timingsEncl} we illustrate the correlation between the timings and the number of tetrahedra contained in the enclosing volume, i.e., those tetrahedra that required to be shrunk. 
This volume is bounded by the Steiner points, however the peaks  that appears in the graphs corresponds to the regions where the trajectory overlaps with a region previously mapped, therefore the number of the tetrahedra contained in the the volume is higher. However the number of enclosed tetrahedra is still $O(1)$.

\begin{table}[t]
\caption{Run times for each step on KITTI sequences: times are expressed  in seconds, frequencies in frame/second}
\label{tab:res}
\centering
\setlength{\tabcolsep}{1px}
\begin{tabular}{lcccccccc}
\toprule 
& & Initialize  &  & Point & Ray & & & \\
& & Steiner & Shrink & Insertion & Tracing & Grow &Total&freq\\
\midrule
\multirow{3}{*}{seq00}
&\cite{litvinov_lhuillier_13} & 0.07& 31.59 & 4.23 & 41.49&257.74&335.12&2.98\\
&\cite{romanoni15b}  & 0.07 & 54.68 &3.95 & 39.21&154.18 &252.09&3.97\\
&proposed  &\textbf{0.03} & \textbf{3.06}& \textbf{0.82} & \textbf{4.50}&\textbf{4.66}&\textbf{13.07}&\textbf{76.51}\\
\midrule
\multirow{3}{*}{seq01}
&\cite{litvinov_lhuillier_13} & 1.21 & 76.78 & 5.12 & 50.60 & 1165.37 & 1299.19 & 0.85\\
&\cite{romanoni15b}& 1.32 & 112.08 & 6.09 & 56.50 & 945.81 & 1121.8 & 0.98\\
&proposed & \textbf{0.84 } & \textbf{7.23} & \textbf{1.83} & \textbf{7.78} & \textbf{6.6} & \textbf{24.28} & \textbf{45.35}\\
\midrule
\multirow{3}{*}{seq02}
&\cite{litvinov_lhuillier_13} & 0.56 & 144.60 & 11.17 & 116.16 & 1957.12 &2229.61 & 0.90\\
&\cite{romanoni15b} & 0.38 & 239.51 & 13.00  & 137.89 & 1936.38 & 2327.16 & 0.86\\
&proposed & \textbf{0.19} & \textbf{13.53} & \textbf{4.16} & \textbf{18.51} & \textbf{11.35} & \textbf{47.74} & \textbf{41.89} \\
\midrule
\multirow{3}{*}{seq05}
&\cite{litvinov_lhuillier_13} & 0.10 & 84.14 & 6.50 & 58.69 & 480.83 & 630.26 & 2.06\\
&\cite{romanoni15b} & 0.07 & 181.79 & 8.93 & 76.42 & 386.72 & 653.93 & 1.99\\
&proposed  & \textbf{0.03} & \textbf{8.46} & \textbf{1.95} & \textbf{8.19} & \textbf{11.99} & \textbf{30.62} & \textbf{42.46}\\
\end{tabular}
\end{table}

\begin{figure}[tp]
\centering
\setlength{\tabcolsep}{1px}
 \begin{tabular}{cc}
  \includegraphics[height=0.28\columnwidth,width=0.44\columnwidth]{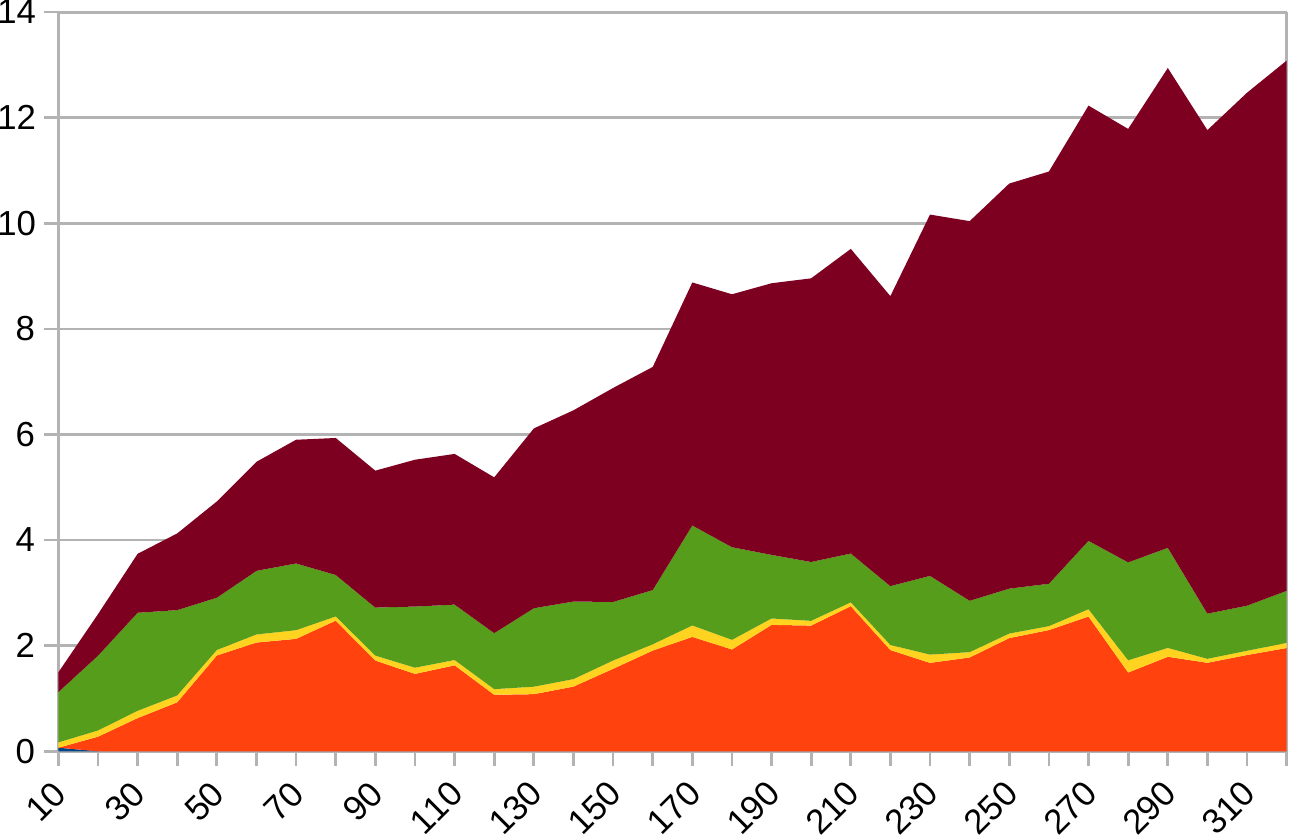}&
  \includegraphics[height=0.28\columnwidth,width=0.44\columnwidth]{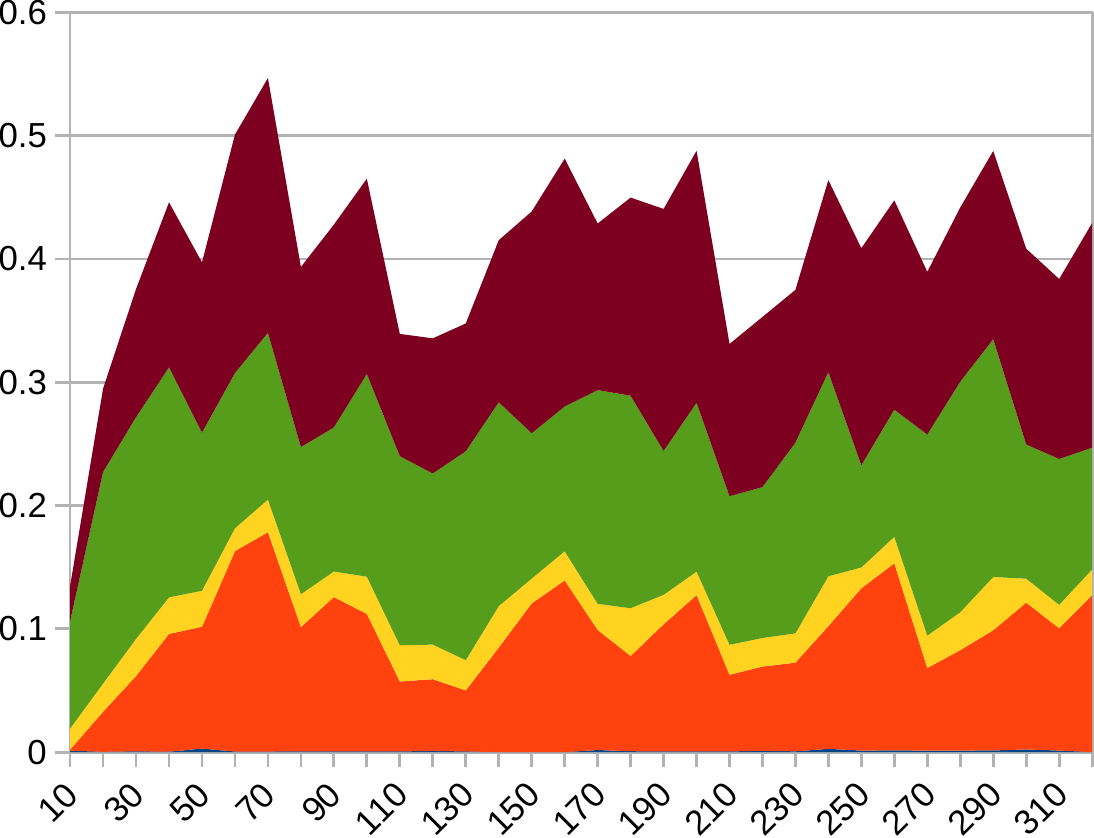}\\
  \multicolumn{2}{c}{sequence 00}\\
  \includegraphics[height=0.28\columnwidth,width=0.44\columnwidth]{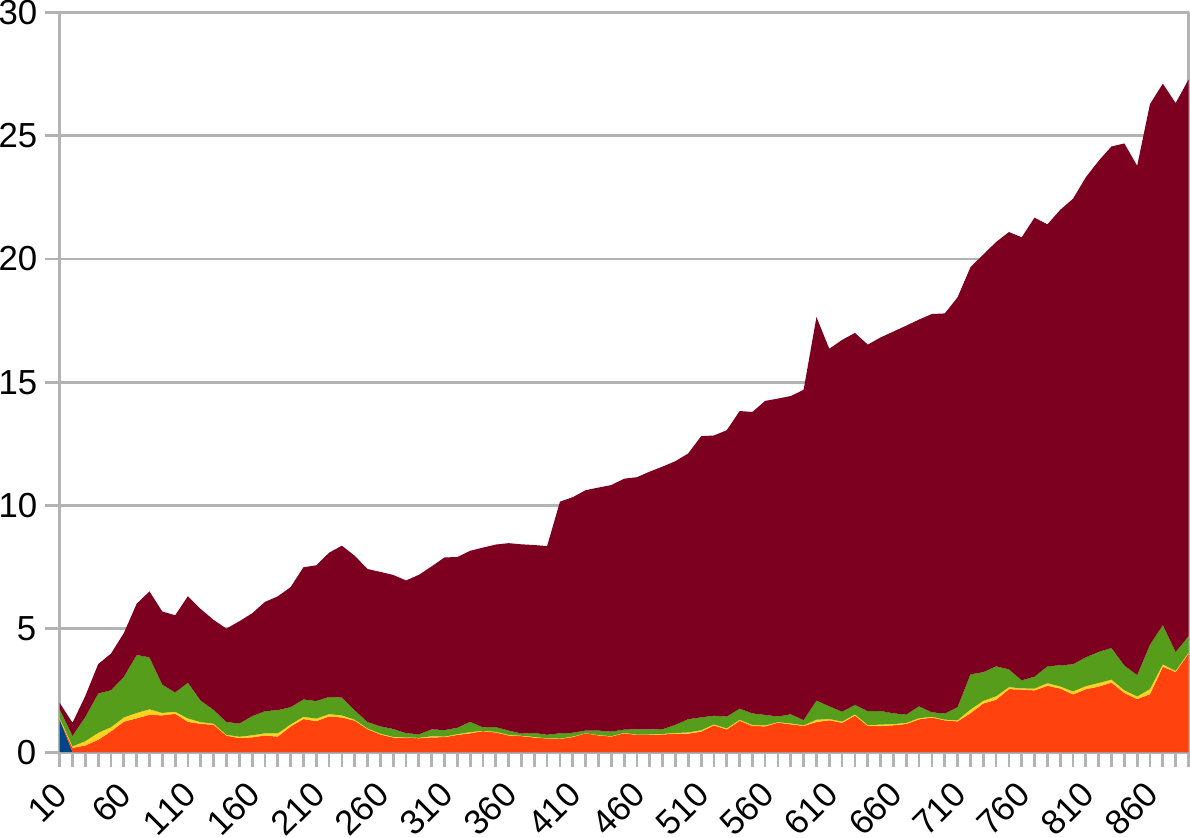}&
  \includegraphics[height=0.28\columnwidth,width=0.44\columnwidth]{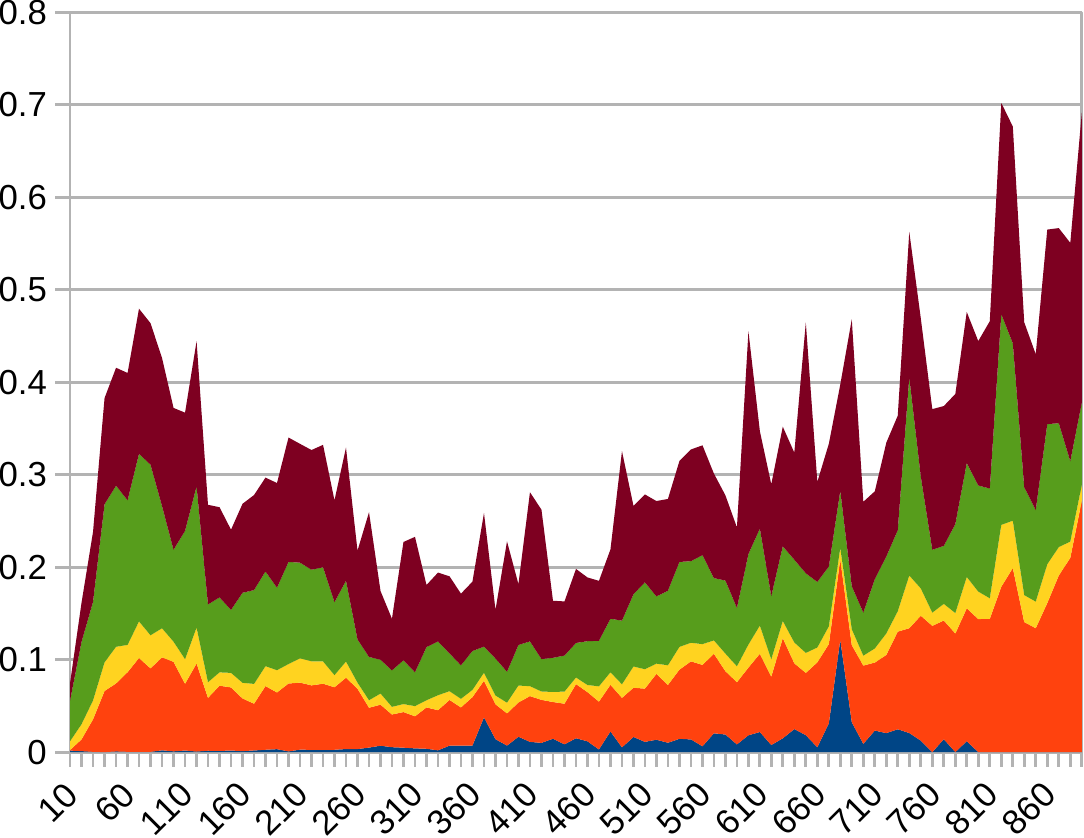}\\
  \multicolumn{2}{c}{sequence 01}\\
  \includegraphics[height=0.28\columnwidth,width=0.44\columnwidth]{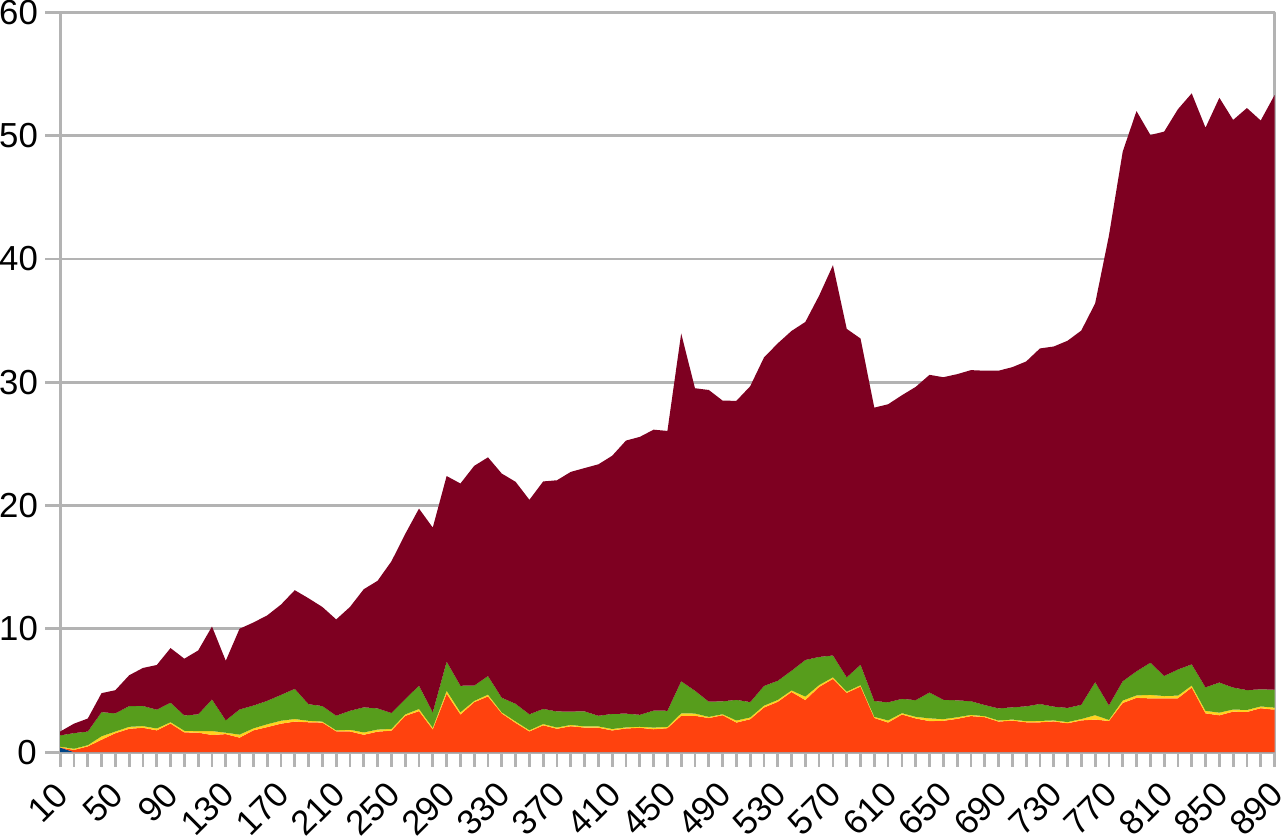}&
  \includegraphics[height=0.28\columnwidth,width=0.44\columnwidth]{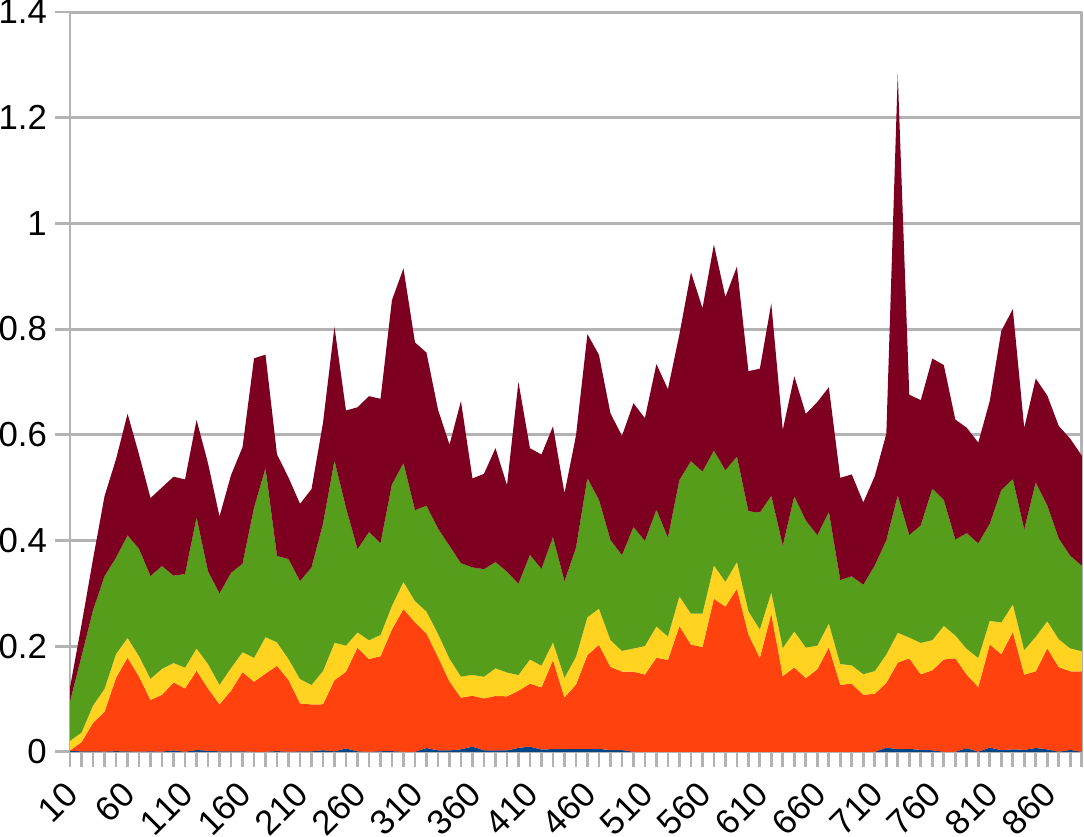}\\
  \multicolumn{2}{c}{sequence 02}\\
  \includegraphics[height=0.28\columnwidth,width=0.44\columnwidth]{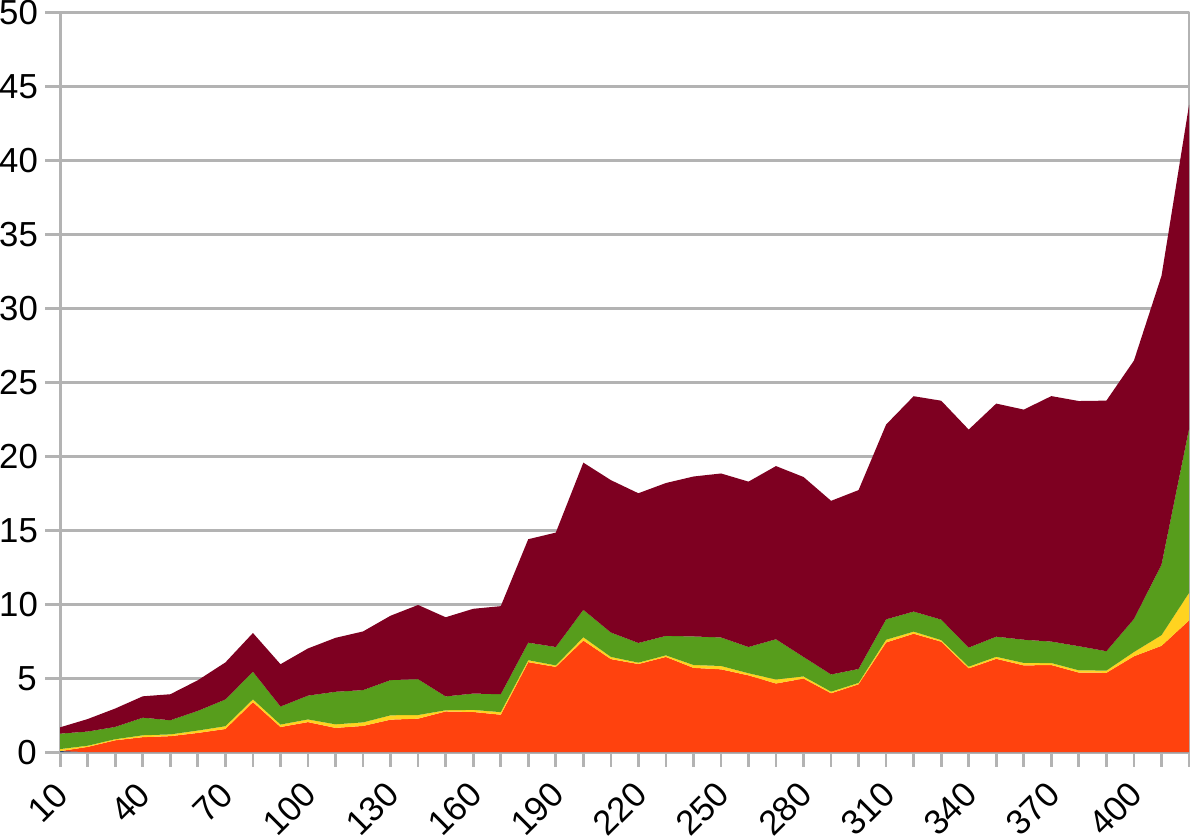}&
  \includegraphics[height=0.28\columnwidth,width=0.44\columnwidth]{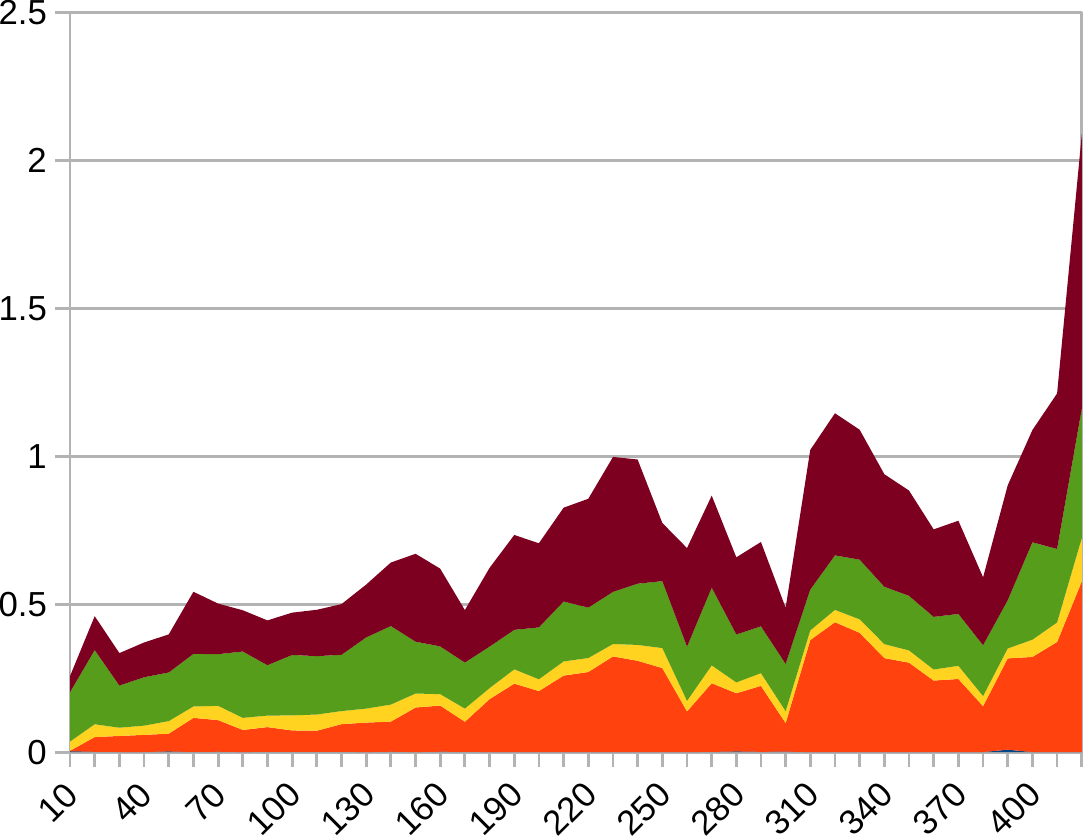}\\
  \multicolumn{2}{c}{sequence 05}\\
  \multicolumn{2}{c}{\includegraphics[width=0.9\columnwidth]{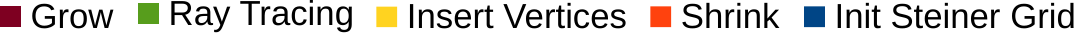}}
 \end{tabular}
 \caption{Timings for each step: on the left~\cite{romanoni15b}; on the right the proposed algorithm (x-axis number of keyframe, y-axis: timings in seconds).}
 \label{fig:timings}
\end{figure}

\begin{figure}[tp]
\centering
\setlength{\tabcolsep}{1px}
 \begin{tabular}{cc}
  \includegraphics[height=0.30\columnwidth,width=0.44\columnwidth]{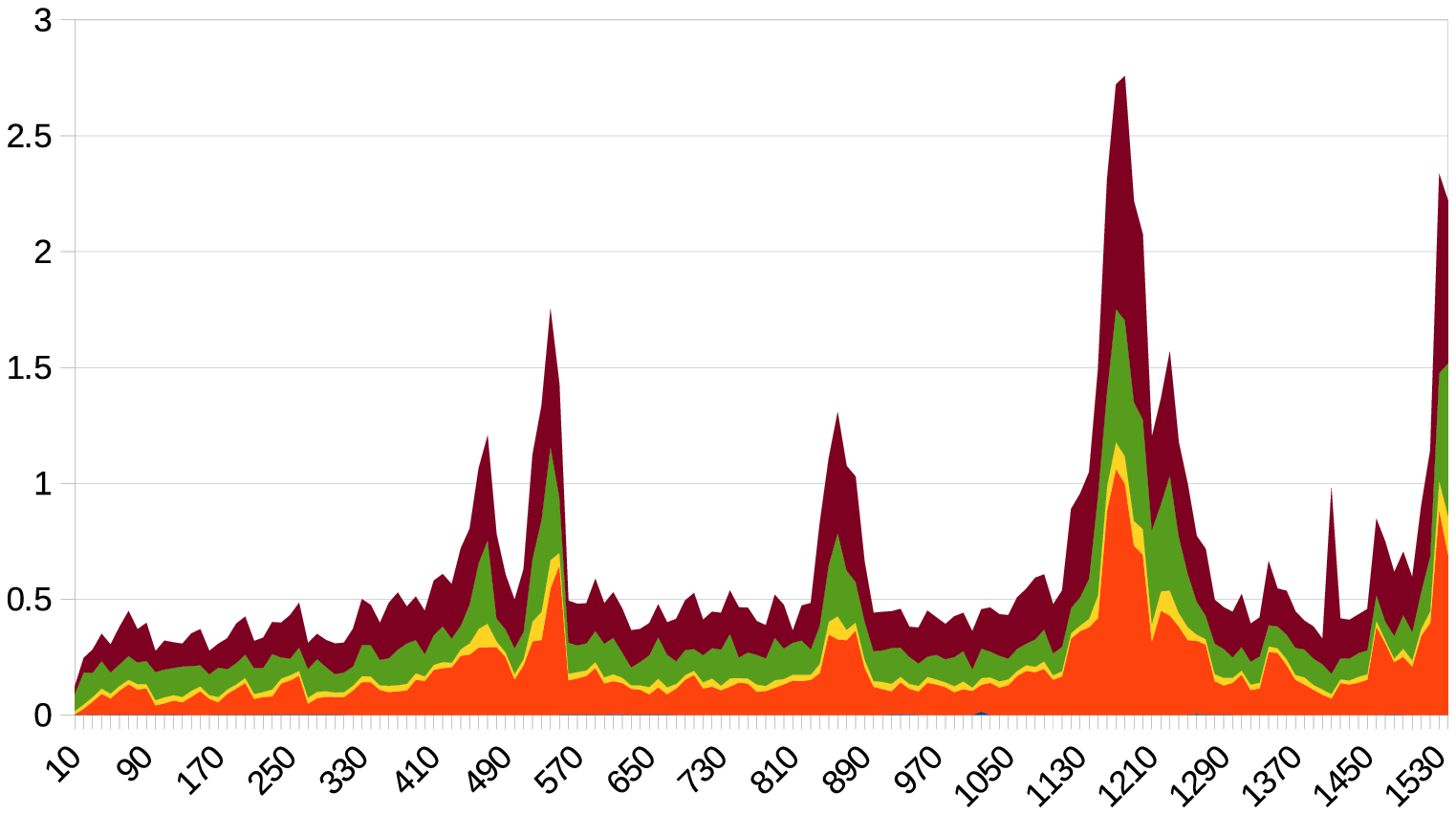}&
  \includegraphics[height=0.30\columnwidth,width=0.44\columnwidth]{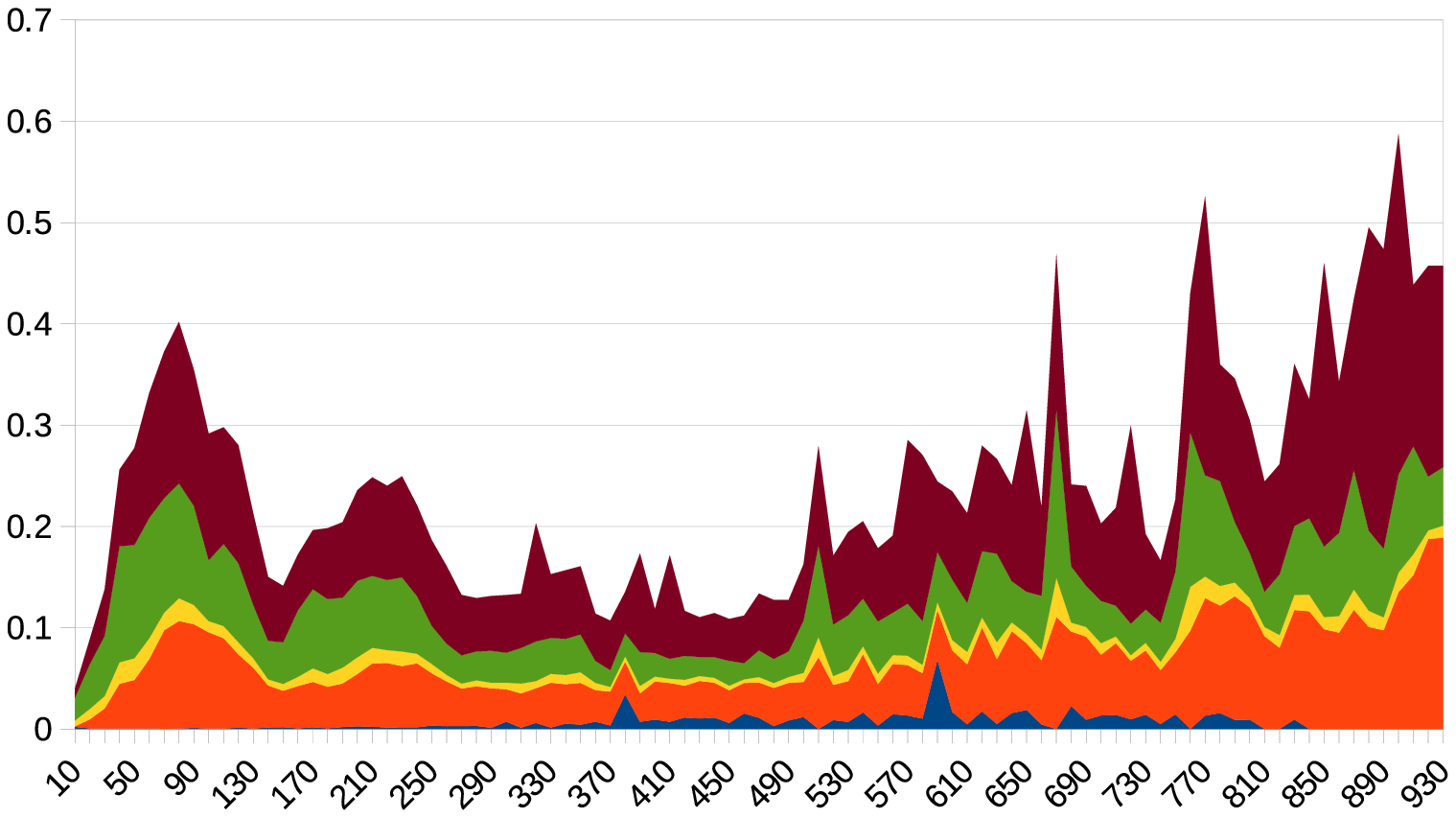}\\
  sequence 00&
  sequence 01\\
  \includegraphics[height=0.30\columnwidth,width=0.44\columnwidth]{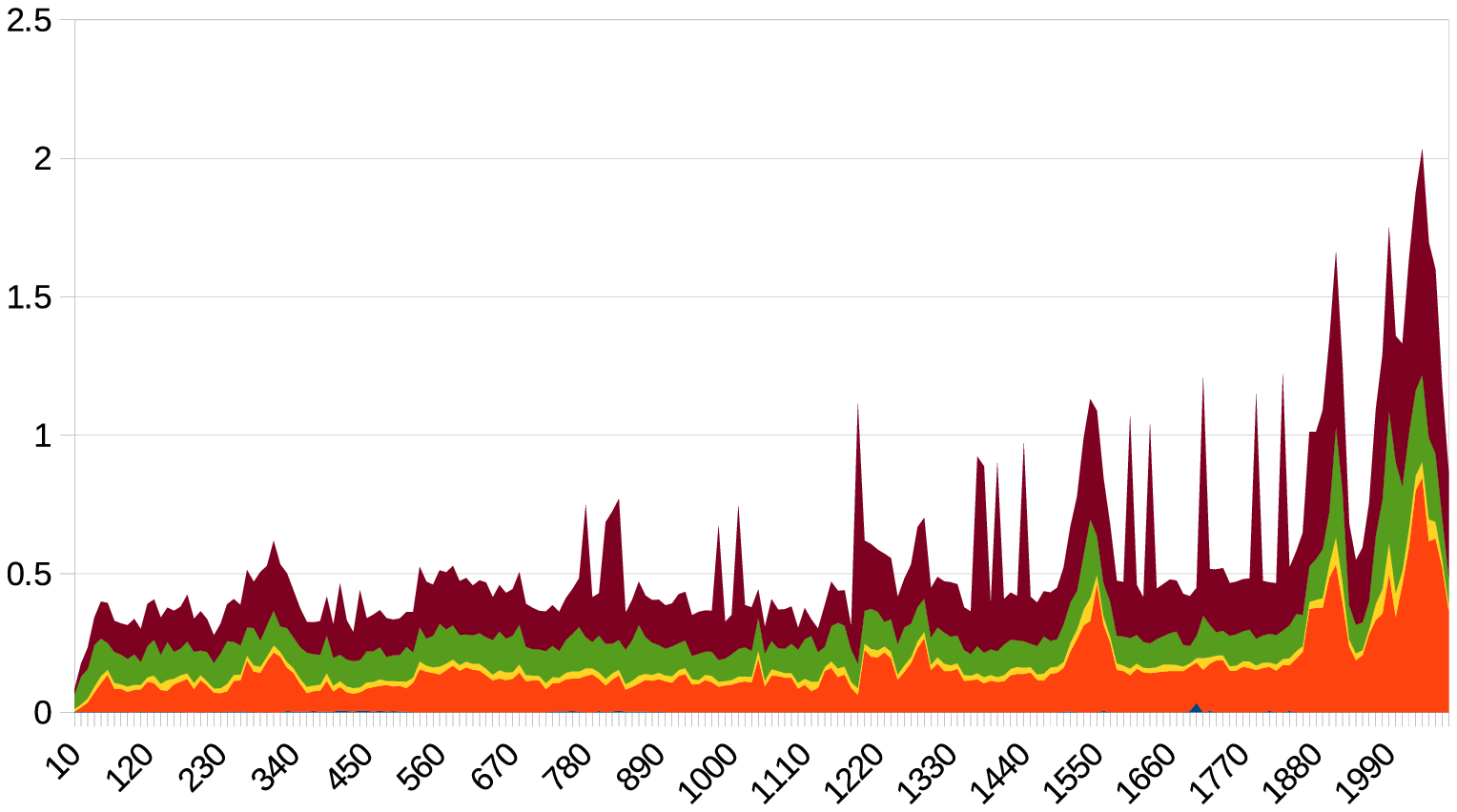}&
  \includegraphics[height=0.30\columnwidth,width=0.44\columnwidth]{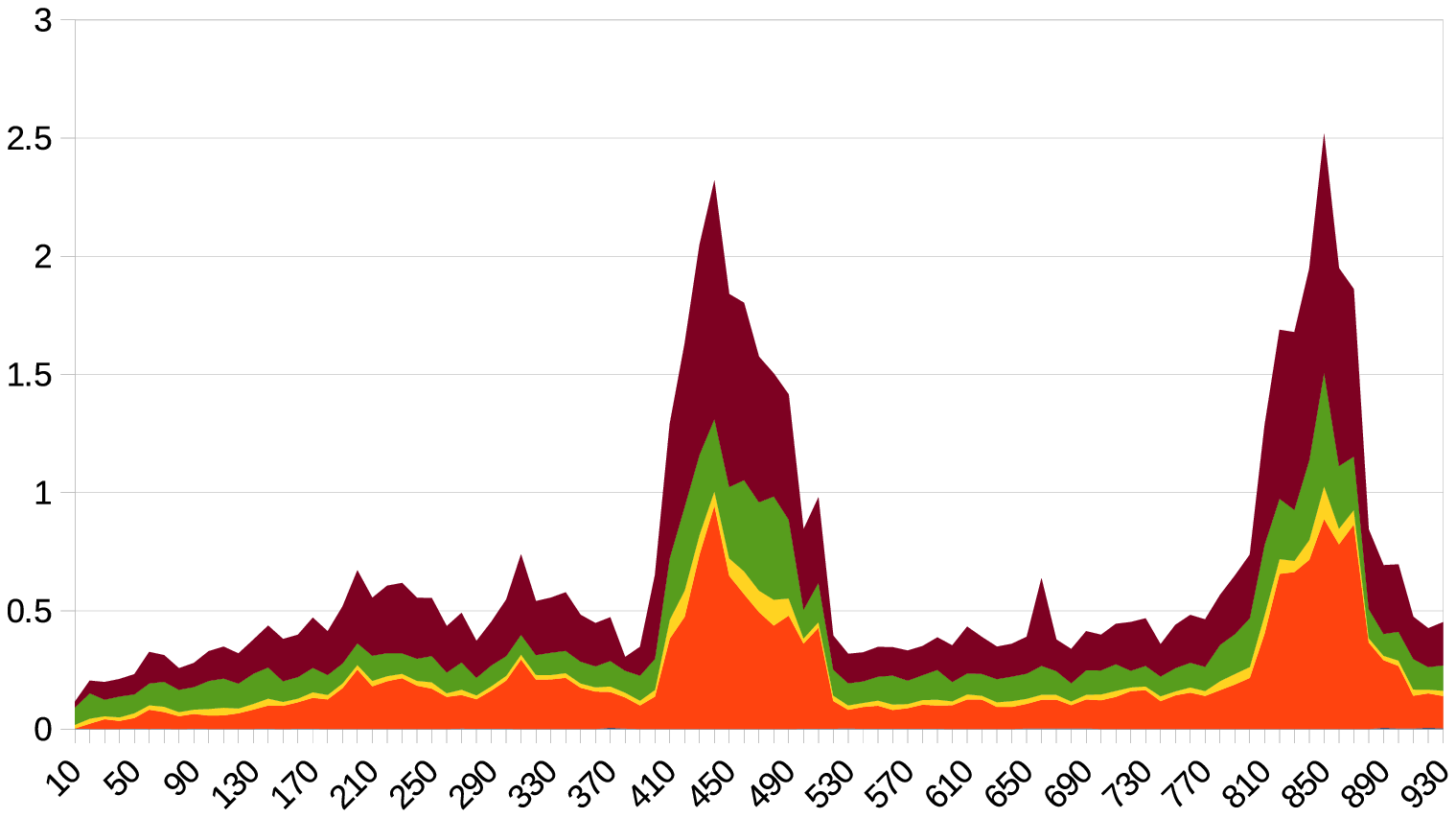}\\
  sequence 02&
  sequence 05\\
  \multicolumn{2}{c}{\includegraphics[width=0.9\columnwidth]{legend}}
 \end{tabular}
 \caption{Timings for each step from the proposed algorithm on the complete sequences (x-axis number of keyframe, y-axis: timings in seconds).}
 \label{fig:timings-complete-sequences}
\end{figure}

\begin{figure*}[t]
\centering
 \begin{tabular}{cccc}
  \includegraphics[width=0.44\columnwidth]{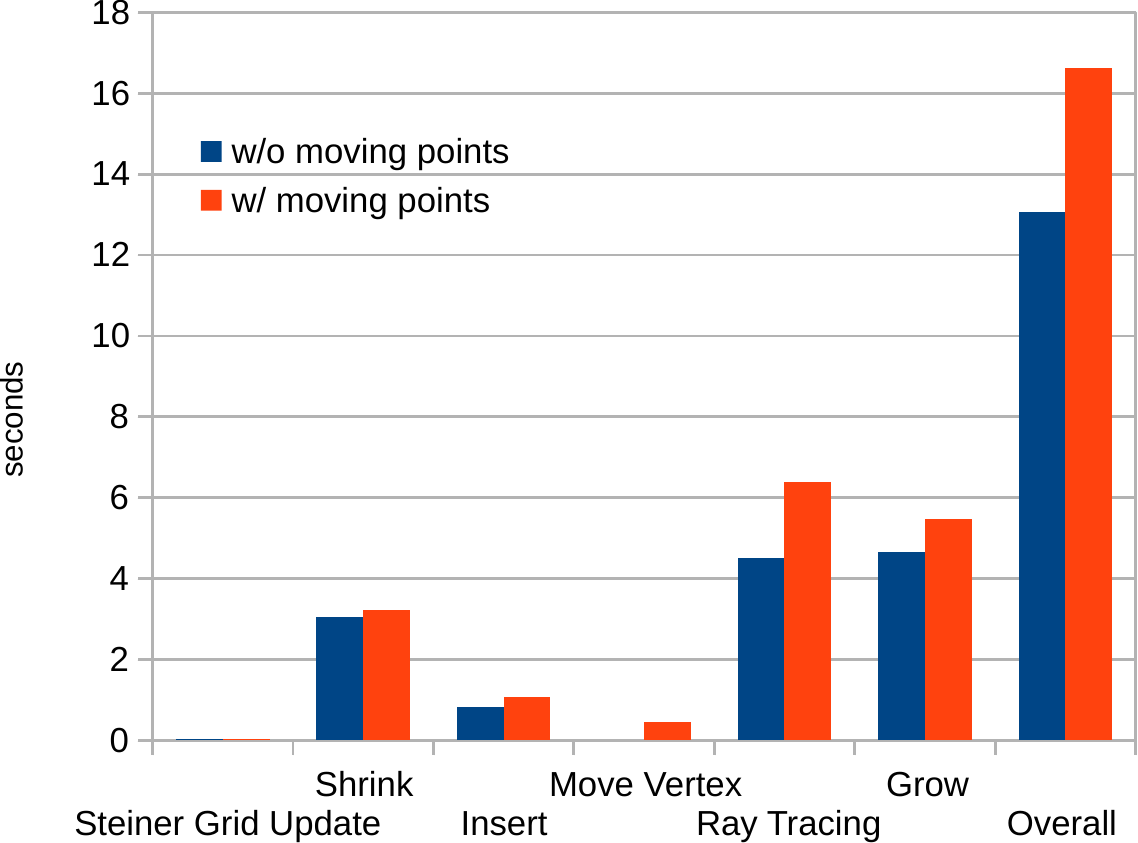}&
  \includegraphics[width=0.44\columnwidth]{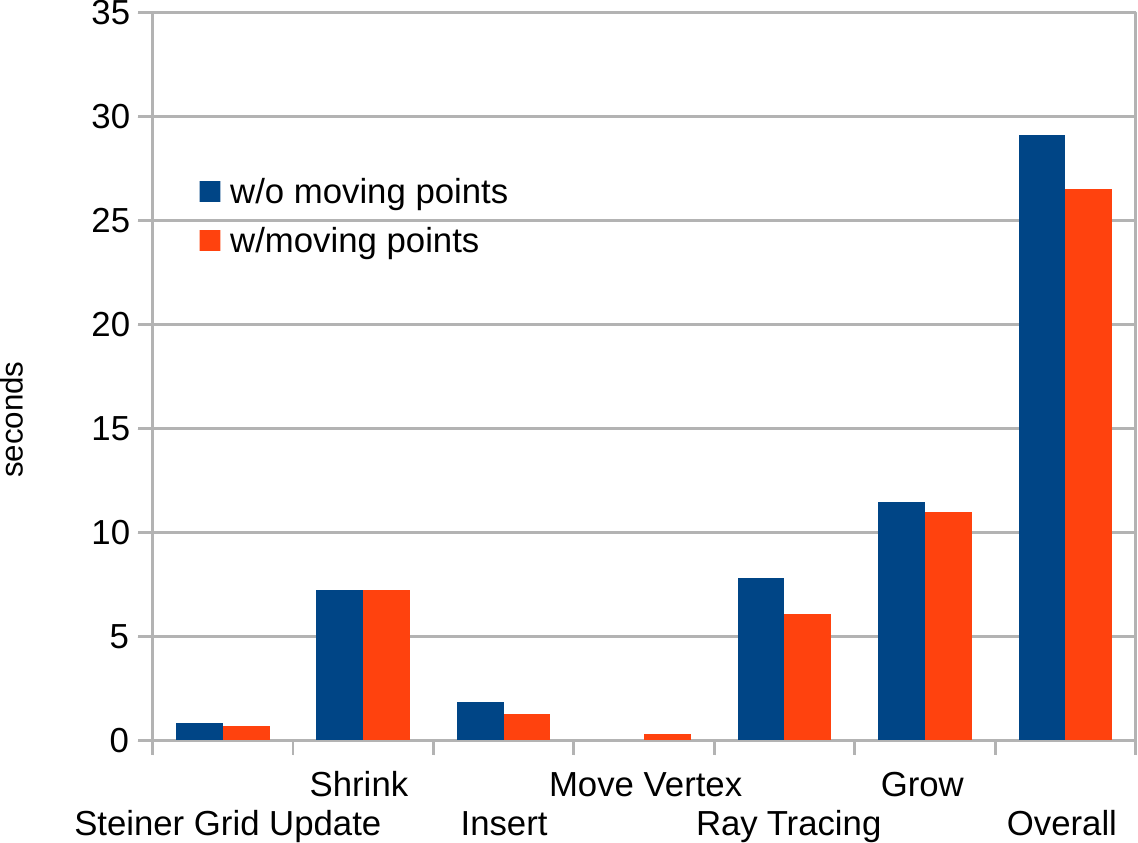}&
  \includegraphics[width=0.44\columnwidth]{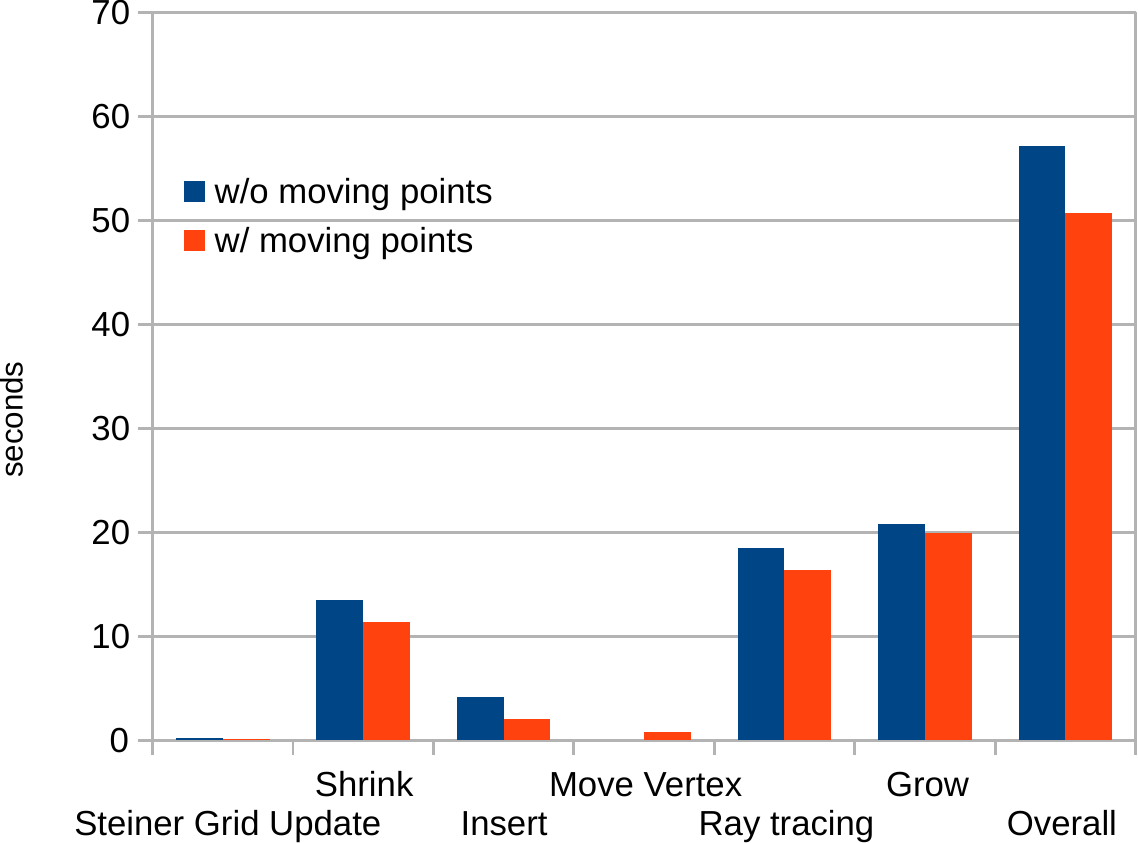}&
  \includegraphics[width=0.44\columnwidth]{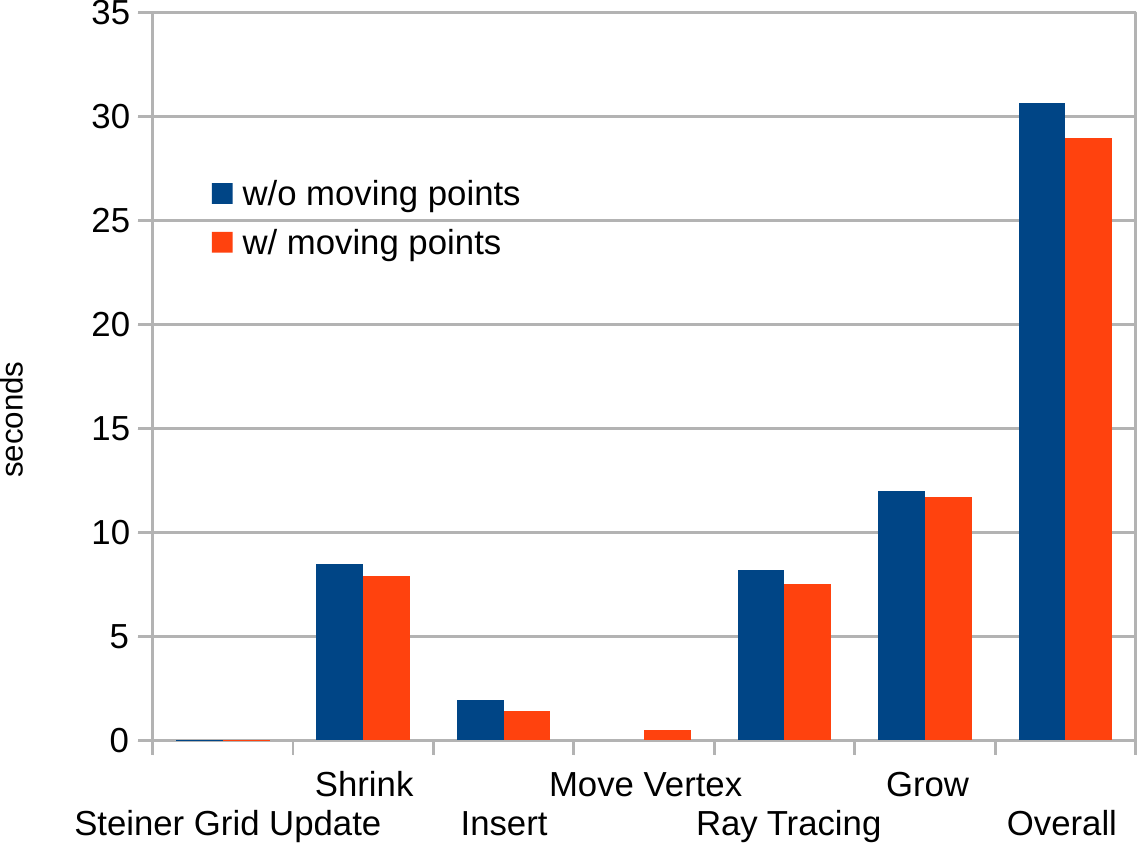}\\
  Sequence 00&
  Sequence 01&
  Sequence 02&
  Sequence 05\\
 \end{tabular}
 \caption{Comparisons of run times with and without the moving points.}
 \label{fig:timingsMoving}
\end{figure*}

We tested also the accuracy of the reconstruction by comparing the depth images of the reconstruction rendered in each frame, against the distance of the velodyne points projected on the same image. In this case we use the velodyne data as ground truth for the 3D reconstructed mesh.
In addition to real-time performance, our algorithm also results in a more accurate reconstruction with respect both to~\cite{romanoni15b} and~\cite{litvinov_lhuillier_13} (see Table~\ref{tab:resacc}). 
This is mainly because the shrink heuristic avoids shrinking big parts of meshes which could remain shrunk after the successive growing process to preserve manifoldness.
Overall the experiments show that the ORB-SLAM point cloud leads to similar results with respect to the Harris-based Structure from motion point cloud.

Since both~\cite{romanoni15b} and~\cite{litvinov_lhuillier_13} are not able to manage moving points, in the previous experiments we have integrated the points positions after they have been observed by two keyframes, and then they are not changed anymore.
We have also evaluated the performance of our algorithm in presence of moving points, i.e., when ORB-SLAM updates the estimate of a point already in the mesh we change its position in the triangulation and we update its visibility information;  even if the presence of moving points more likely causes the shrink procedure to get stuck in local minima and the generation of small visual artifacts, the resulting accuracy is not significantly diminished and still remains comparable with respect to both~\cite{romanoni15b} and~\cite{litvinov_lhuillier_13}.

In Fig.~\ref{fig:timingsMoving} we show how the moving points handling affects the run times. The points which moved from their first insertion in the triangulation correspond to the 9\%, 3\%, 7\% and 8\% respectively in the 00, 01, 02, 05 sequences.
In Table \ref{tab:stats} we reported the complete average per keyframe statistics. 
Since these points are usually very close to the vertex we need to modify, the computational overhead is marginal: indeed, Figure~\ref{fig:timingsMoving} shows very similar timings and variations are only to be ascribed to experiment-specific latencies.

\begin{table}[t]
\caption{Mean Absolute Error (MAE) of the reconstruction for the KITTI sequences errors in meters}
\label{tab:resacc}
\centering
\begin{tabular}{lcccc}
\toprule 
&  seq00 & seq01 & seq02 & seq05\\
\midrule
\cite{litvinov_lhuillier_13} (w/o moving points) & 1.32 & 0.79 & 0.70 & 1.48 \\
\cite{romanoni15b}  (w/o moving points) & 1.13 & 1.22 & 0.72 & 1.50  \\
proposed  (w/o moving points) & \textbf{0.91} & \textbf{0.72} & \textbf{0.62} & \textbf{1.14} \\
proposed (w/ moving points)  & 1.05 & 1.14 & 0.74 & 1.21\\
\end{tabular}
\end{table}


\begin{table}[t]
\caption{Reconstruction average per-keyframe statistics with moving points}
\label{tab:stats}
\centering
\begin{tabular}{lcccccccccccc}
\toprule 
					& seq00 & seq01 & seq02 & seq05 \\
\midrule
\# moving points 	& 12 	& 28 	& 10 	& 10	\\
\# new points 		& 128 	& 72 	& 125 	& 12	\\
\# tetrahedra shrinked 	& 1171 & 739  & 12334 & 2100 \\
\# Steiner cells shrinked	& 26   & 29  & 350 & 27 \\
\# points not added & 0 	& 1     & 0		& 0		\\
\# traced rays 		& 492	& 2188	& 454	& 443	\\
\# re-traced rays 	& 112	& 696	& 97	& 155	\\
\# untraced rays 	& 53	& 110	& 44	& 46	\\
\end{tabular}
\end{table}

\section{CONCLUSION AND FUTURE WORKS}
\label{sec:concl}
We proposed the first real-time incremental manifold reconstruction algorithm that runs on a single CPU and can handle large scale scenarios; the proposed algorithm is able to speed up the run time of state of the art algorithms significantly and, at the same time, it also improves on the accuracy of the reconstruction.
We were able to obtain such results by redesigning few of the classical manifold reconstruction steps proposing a novel shrinking method and a novel ray tracing approach which leverage on hashing and caching strategies.
In contrast to existing algorithm our algorithm is also able to manage moving points without any approximate heuristics and with negligible overheads.

As a possible future development we plan to manage loop closures and update the map shape accordingly, which means handling also the change the mesh genus when needed. We are also working on improving the accuracy of the reconstruction by incorporating shape prior, such as planes, through constrained Delaunay triangulation, still preserving real-time execution.

\section*{ACKNOWLEDGMENT}
{\small
This work has been supported by the POLISOCIAL Grant ``Maps for Easy Paths (MEP)'' and the ``Cloud4Drones'' project founded by EIT Digital. We thank Nvidia who has kindly supported our research through the Hardware Grant Program. }
\bibliographystyle{IEEEtran}
\bibliography{biblioTotal}

\end{document}